\pdfoutput=1

\documentclass[11pt]{article}

\usepackage[preprint]{acl}

\usepackage{times}
\usepackage{latexsym}
\usepackage{algorithm}
\usepackage[noend]{algpseudocode}
\usepackage{amsmath}
\usepackage{amssymb}
\usepackage{booktabs}
\usepackage{enumerate}
\usepackage{colortbl,xcolor}
\usepackage{subfigure}
\usepackage{multirow}
\usepackage{pgfplots}
\usepackage{caption}
\usepackage{tcolorbox}
\tcbuselibrary{most}
\usepackage{arydshln}
\usepackage{float}
\usepackage{pgfplots}
\pgfplotsset{compat=1.17}
\usepackage{sansmath}
\usepackage{pgf-pie}

\usepackage{alltt}

\usepackage[T1]{fontenc}

\usepackage[utf8]{inputenc}

\usepackage{microtype}

\usepackage{inconsolata}

\usepackage{graphicx}

\title{\textit{ConsistentChat}: Building Skeleton-Guided Consistent Multi-Turn \\ Dialogues for Large Language Models from Scratch}

\author{
  \textbf{Jiawei Chen}${}^{1,2}$\thanks{~ Equal contribution}, 
  \textbf{Xinyan Guan}${}^{1,2}$\footnotemark[1],
  \textbf{Qianhao Yuan}${}^{1,2}$, 
  \textbf{Guozhao Mo}${}^{1,2}$, \\
  \textbf{Weixiang Zhou${}^{1}$}\thanks{~ Corresponding authors},
  \textbf{Yaojie Lu${}^{1}$},
  \textbf{Hongyu Lin${}^{1}$},
  \textbf{Ben He${}^{1,2}$\footnotemark[2]},
  \textbf{Le Sun${}^{1}$},
  \textbf{Xianpei Han${}^{1}$}
  \\
  ${}^{1}$Chinese Information Processing Laboratory, Institute of Software, \\
  Chinese Academy of Sciences, Beijing, China \\
  ${}^{2}$University of Chinese Academy of Sciences, Beijing, China \\
 \texttt{\{chenjiawei2024,guanxinyan2022,yuanqianhao2024,moguozhao2024\}@iscas.ac.cn}\\
 \texttt{\{weixiang,luyaojie,hongyu,sunle,xianpei\}@iscas.ac.cn} ~
 \texttt{benhe@ucas.ac.cn}
}

\begin{document}
\maketitle
\begin{abstract}
Current instruction data synthesis methods primarily focus on single-turn instructions and often neglect cross-turn coherence, resulting in context drift and reduced task completion rates in extended conversations. To address this limitation, we propose Skeleton-Guided Multi-Turn Dialogue Generation, a framework that constrains multi-turn instruction synthesis by explicitly modeling human conversational intent. It operates in two stages: (1) Intent Modeling, which captures the global structure of human dialogues by assigning each conversation to one of nine well-defined intent trajectories, ensuring a coherent and goal-oriented information flow; and (2) Skeleton Generation, which constructs a structurally grounded sequence of user queries aligned with the modeled intent, thereby serving as a scaffold that constrains and guides the downstream instruction synthesis process. Based on this process, we construct \textit{ConsistentChat}\footnote{Our code, model and dataset are publicly available at \url{https://github.com/chenjiawei30/ConsistentChat}}, a multi-turn instruction dataset with approximately 15,000 multi-turn conversations and 224,392 utterances. Experiments on the \textsc{Light}, \textsc{TopDial}, and \textsc{MT-Eval} benchmarks show that models fine-tuned on \textit{ConsistentChat} achieve a 20–30\% improvement in consistency and up to a 15\% increase in task success rate, significantly outperforming models trained on existing single-turn and multi-turn instruction datasets.
\end{abstract}

\section{Introduction}

\begin{figure}[t!]
\centering
\subfigure[Single-Turn Instruction Synthesis Manners.]{
    \begin{minipage}[b]{0.93\linewidth}
	    \includegraphics[width=1\linewidth]{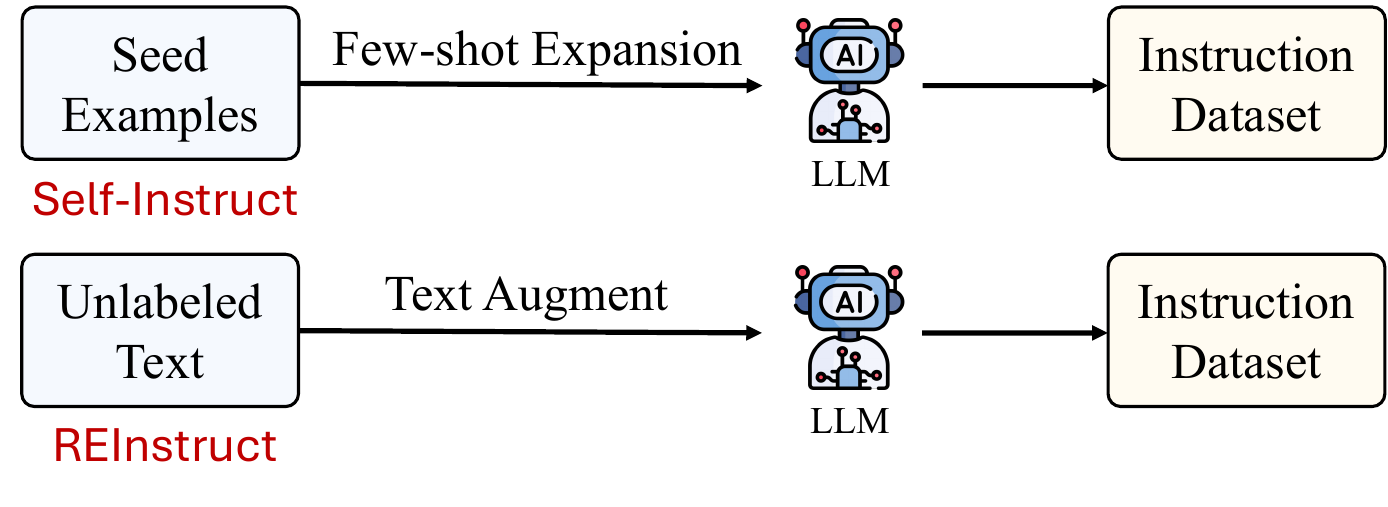}
    \end{minipage}
    \label{fig:paradigm_single}
}
\subfigure[Multi-Turn Instruction Synthesis Manners.]{
    \begin{minipage}[b]{0.975\linewidth}
	    \includegraphics[width=1\linewidth]{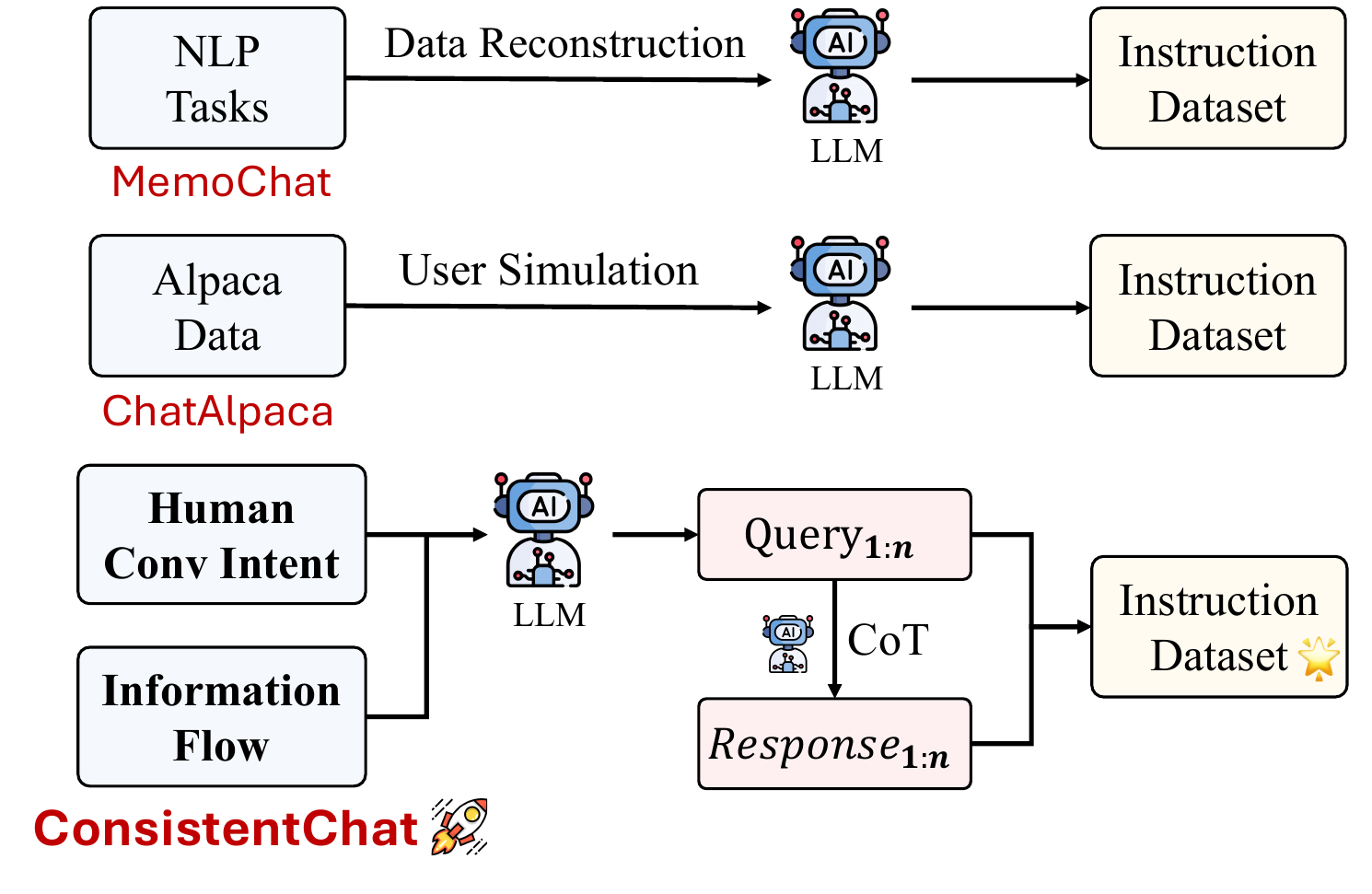}
    \end{minipage}
    \label{fig:paradigm_multi}
}
\caption{Comparison of single-turn and multi-turn instruction synthesis manners.}
\label{fig:intro}
\end{figure}

Instruction synthesis plays a fundamental role in enabling large language models (LLMs) to perform a wide array of real-world dialogue tasks \cite{iyer2022opt}. High-quality instruction datasets are indispensable for supervised fine-tuning (SFT), allowing LLMs to better capture user intent and handle diverse conversational scenarios, such as customer support and educational tutoring \cite{sanh2022multitask, wang2023self}. As the demand for scalable and domain-adaptive models grows, the automatic construction of instruction data has become a challenge in the development of advanced dialogue systems.

Currently, existing instruction data construction paradigms are primarily tailored to single-turn interactions. Works such as Self-Instruct \cite{wang2023self}, Evolve-Instruct \cite{xu2023wizardlm}, and REInstruct \cite{chen2024reinstruct} leverage prompt expansion and corpus mining to generate large-scale <instruction, response> pairs. However, these methods mainly focus on turn-level exchanges, overlooking the sequential dependencies and interactive dynamics intrinsic to natural conversations. This gap not only limits the capacity of models to capture real conversational development, but also fails to reflect the actual requirements of real-world applications, where multi-turn interactions are the norm rather than the exception.

Fundamentally, some approaches explored the synthesis of multi-turn instruction datasets, such as MemoChat \cite{lu2023memochat}, ChatAlpaca \cite{ChatAlpaca} and GLAN \cite{li2024synthetic}. These approaches largely rely on either transforming traditional NLP tasks into dialogues or simulating multi-turn conversations with LLMs. Such strategies are typically unconstrained: the generation process focuses on producing locally coherent, turn-level instruction with response pairs, failing to model the evolution of human intent or the overall dialogue trajectory. As a result, these datasets often suffer from poor chat consistency across turns, with frequent occurrences of topic drift throughout the conversation. This limitation poses a challenge for training models to maintain alignment and consistency in multi-turn interactions.

Based on the above observations, we propose an instruction synthesis framework that can model human conversational intent and information flow to guide multi-turn dialogue generation and consequently construct the \textit{ConsistentChat} dataset. We analyze nine categories of dialogue intent \cite{rapp2021human} (e.g., Problem-Solving-Interaction and Educational-Interaction) and formalize them as dynamic information flows that constrain both topic progression and role interaction throughout the dialogue. Our approach first generates a globally coherent sequence of user queries conditioned on the underlying intent, and subsequently produces agent responses that maintain alignment with both current and global conversational objectives.

We evaluate models trained with \textit{ConsistentChat} on \textsc{Light}, \textsc{TopDial}, and \textsc{MT-Eval} benchmarks. Preliminary analyses (\S\ref{sec:preliminary_analysis}) reveal that LLMs’ multi-turn dialogue abilities degrade with increasing dialogue depth, and strong chat consistency in training data is crucial for SFT models' alignment across turns. Leveraging our Skeleton-Guided framework, we build high-quality multi-turn instruction dialogues. Experimental results show that models fine-tuned on \textit{ConsistentChat} instruction consistently outperform those trained on existing datasets, both in single-turn and multi-turn settings, establishing state-of-the-art results in dialogue chat consistency and alignment. Our findings highlight the importance of modeling human conversational intent and information flow for generating reliable instruction data for multi-turn dialogue agents.

The main contributions of this paper are:

1) This is the first work investigating how modeling human conversational intent in instruction synthesis influences the chat consistency of fine-tuned dialogue models. 

2) We propose \textit{ConsistentChat}, generated by a simple yet effective Skeleton-Guided framework for supervised fine-tuning, which can be applied in broad downstream dialogue scenarios.

3) Extensive experiments demonstrate that instruction data generated by \textit{ConsistentChat} outperforms existing popular multi-turn datasets in terms of chat consistency, as well as both single-turn and multi-turn conversational capability.

\section{Revealing Multi-Turn Degradation and Chat Consistency Effects}
\label{sec:preliminary_analysis}

In this section, we first investigate the performance of popular dialogue models in multi-turn conversations and explore how fine-tuning on datasets with varying chat consistency affects their performance. We conduct analyses addressing these two questions:

\vspace{-8pt}
\begin{itemize}
\item\textbf{Question I}: When current conversational models engage in multi-turn dialogues, how do the multi-turn conversational abilities evolve as the conversation progresses?
\item\textbf{Question II}: Whether the chat consistency of dialogue datasets affects the performance of fine-tuned models?
\end{itemize}
\vspace{-8pt}

In this paper, we define \textbf{chat consistency} as the semantic coherence of a dialogue across multiple turns. Specifically, it encompasses three key aspects: (1) Alignment with the initial conversational intent, (2) Smooth and logically connected information flow, and (3) Contextual relevance of each response across turns. This definition is closely related to prior work on consistency in dialogue, which emphasizes maintaining alignment in styles and topics \cite{wang-etal-2017-steering}, personas \cite{ju2022learning}, and conversational characters or roles \cite{wang-etal-2024-instruct}.

We conducted experiments on \textsc{MT-Eval} \cite{kwan2024mt} to assess the multi-turn conversational abilities of mainstream dialogue models, addressing Question I (see Appendix \ref{appendix:preliminary_analysis_1} for details). For Question II, we further curated subsets of instructions from \href{https://huggingface.co/datasets/shibing624/sharegpt_gpt4}{\texttt{ShareGPT}} \cite{vicuna2023}, categorized into three levels of chat consistency (Low, Sample, and High). These subsets were then used to fine-tune pre-trained LLaMA-3.1-8B model, evaluated with the chat consistency metric on \textsc{Light} \cite{urbanek-etal-2019-learning} and \textsc{TopDial} \cite{wang-etal-2023-target} (see Appendix \ref{appendix:preliminary_analysis_2}).

\subsection{Conversational Abilities Diminish as the Dialogue Deepens}
\paragraph{Setups.}
We selected several popular dialogue models (including Qwen-2.5, LLaMA-3.1, Mistral-0.3, Vicuna-1.5), covering a total of 8 representative models across different scales and architectures. These models were evaluated on the Refinement task of the \textsc{MT-Eval} benchmark, which measures the ability to generate coherent and contextually relevant responses when conversations extend over multiple turns.

\begin{figure}[t!]
\centering
\includegraphics[width=1\linewidth]{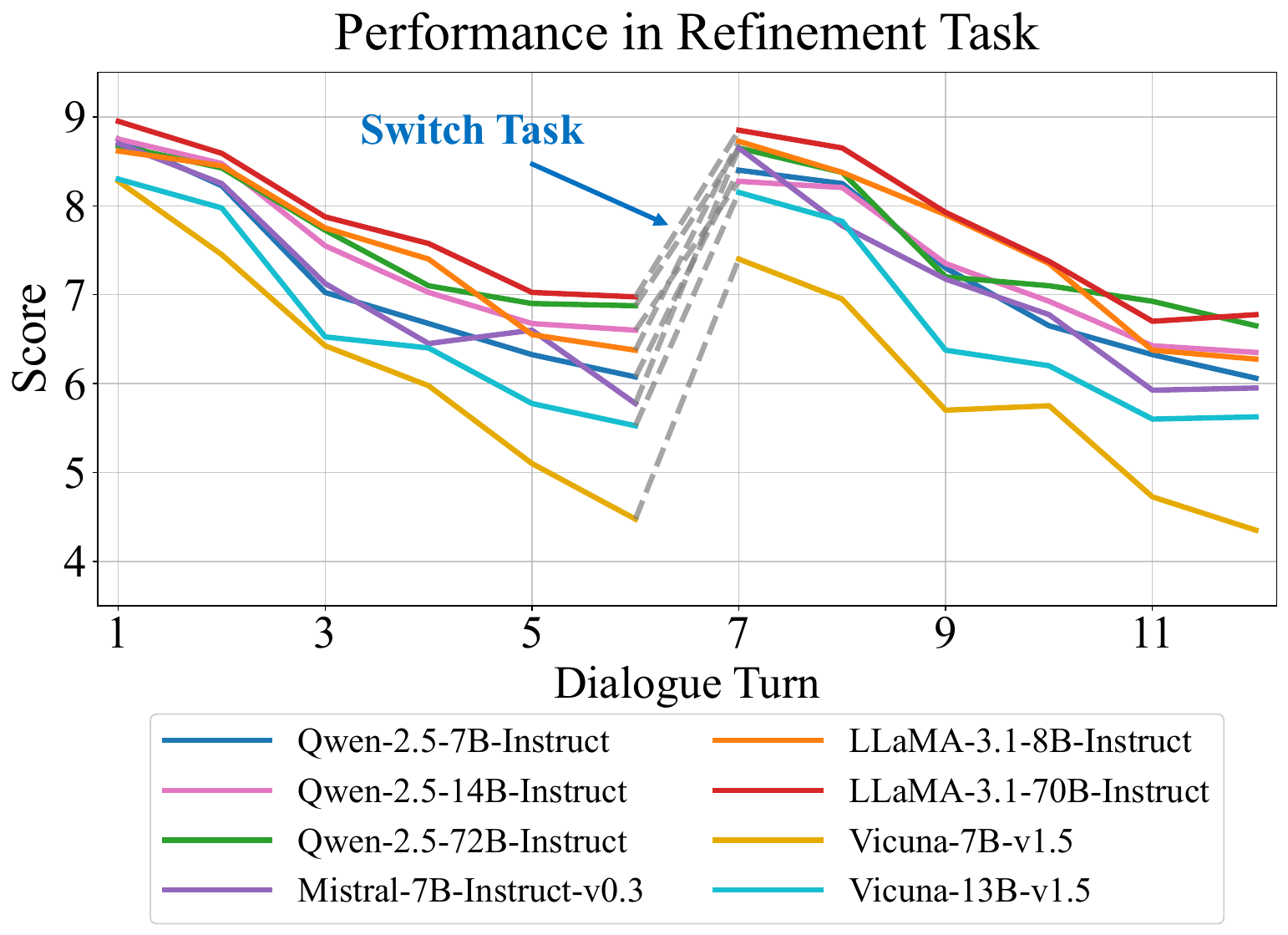}
\caption{Performance of multi-turn conversational ability declines as more instructions are added on Refinement task in \textsc{MT-Eval}. Each dialogue in task has two NLP tasks with each task comprising six increasingly complex instructions. The transition to the second NLP task occurs at the seventh turn as denoted by the gray dashed line. Judged by Qwen-2.5-72B-Instruct.}
\label{fig:pre_fig_1}
\end{figure}

\paragraph{Results.}
Figure \ref{fig:pre_fig_1} shows that as dialogue turns increase, model performance consistently deteriorates. This degradation trend is observed across all model families, although larger-scale models exhibit relatively stronger resilience compared to smaller ones. These findings highlight the inherent challenges in preserving consistency during long interactions. Detailed results are in Appendix \ref{appendix:pre_mt_eval}.

\subsection{Dataset Consistency Affects Fine-tuned Model Performance}
\paragraph{Setups.} 
We utilized Qwen-2.5-72B-Instruct to assign a chat consistency score, ranging from 1 to 10, to each multi-turn dialogue from the ShareGPT (GPT-4 version, a high-quality dataset curated and cleaned by \citet{shibing2023sharegpt}). Based on the assigned scores, we categorized the dialogues into three groups: High consistency (scores between 8 and 10), Low consistency (scores between 4 and 6), and a sample of dialogues, with each group containing approximately 10,000 dialogue instances. Subsequently, we performed supervised fine-tuning of the LLaMA-3.1-8B model on these datasets and evaluated on tasks involving chat consistency and multi-turn conversational capabilities. The scoring prompt is illustrated in Figure \ref{fig:llm_eval_sharegpt}.

\begin{figure}[t!]
\centering
\includegraphics[width=1\linewidth]{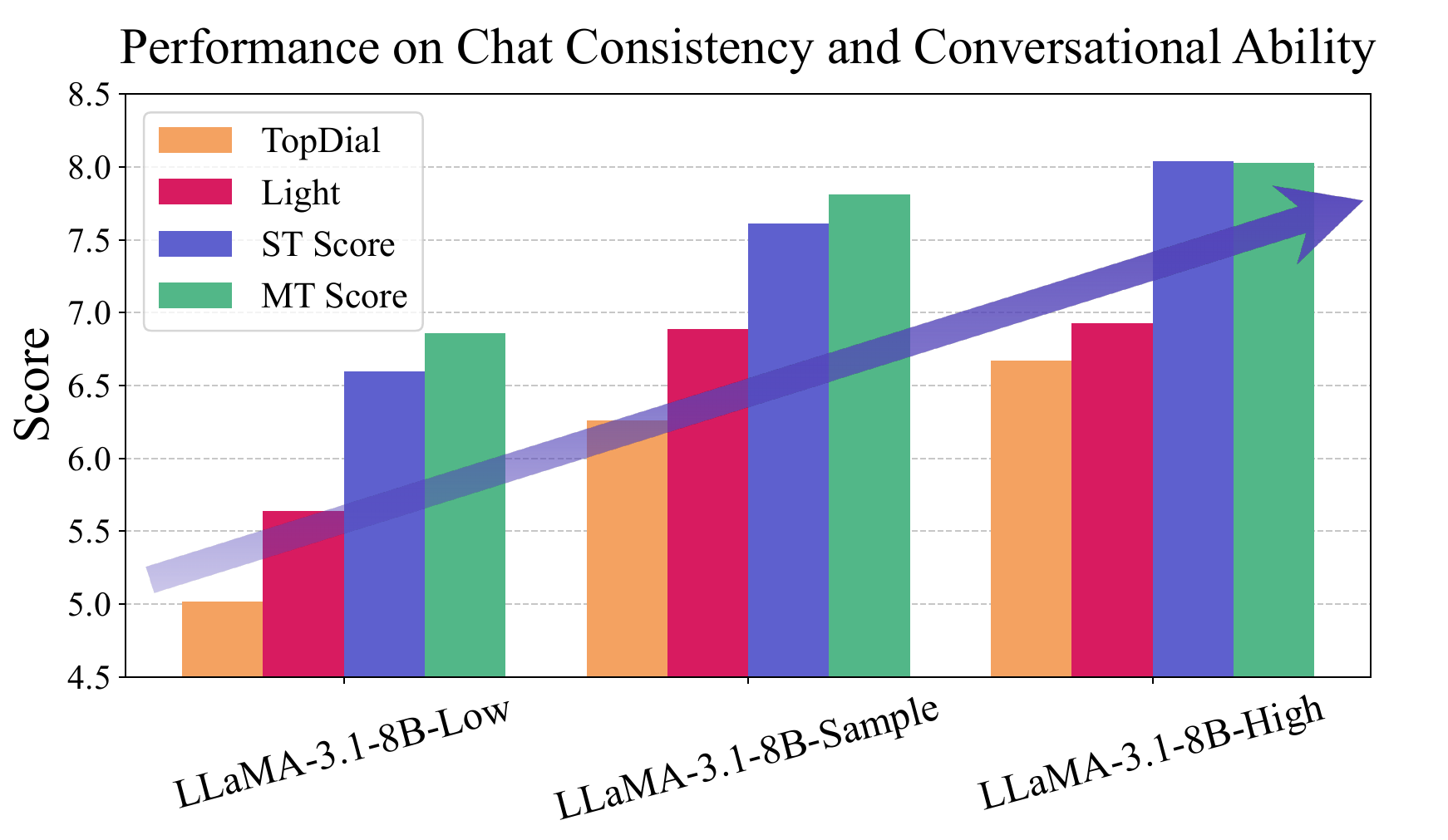}
\caption{Performance of LLaMA-3.1-8B models (Low, Sample, and High consistency) across four evaluation settings: \textsc{TopDial}, \textsc{Light} and Single-Turn (ST), and Multi-Turn (MT) in \textsc{MT-Eval}. The results demonstrate ongoing improvements across all benchmarks as the chat consistency of SFT instructions increases.}
\label{fig:pre_fig_2}
\end{figure}

\paragraph{Results.}
Figure \ref{fig:pre_fig_2} shows that LLaMA-3.1-8B-High achieves the best results in both consistency metrics (detailed in Table \ref{tab:pre_exp_2}) and conversational capabilities (detailed in Table \ref{tab:pre_exp_3}), illustrating that chat consistency plays a crucial role in fine-tuning models' performance. The whole experimental results are shown in Appendix \ref{appendix:preliminary_analysis_2}.

\subsection{Insights from Preliminary Analyses}
Through our preliminary analyses, we have demonstrated that: (1) Popular dialogue models exhibit consistent degradation in conversational abilities as turns increases; (2) The consistency of training data affects the performance of fine-tuned models, with higher consistency data yielding superior results. These findings point out developing high-quality instruction synthesis methods as a promising direction for enhancing multi-turn dialogue models.

\section{Building Consistent Multi-turn Dialogues from Scratch}
\label{sec:exp}

\begin{figure*}[th!]
\centering
\includegraphics[width=1\textwidth]{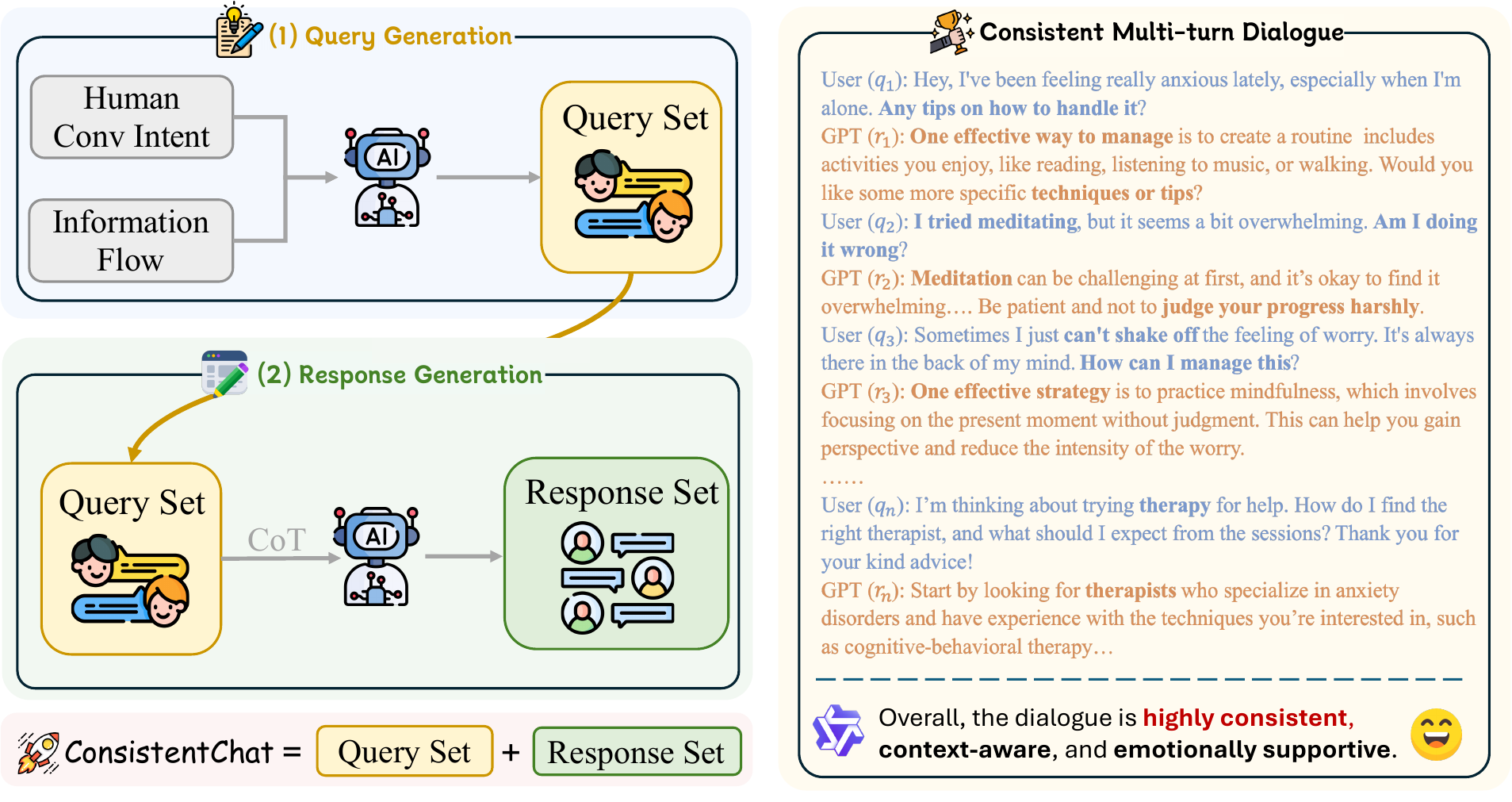}
\caption{Overview of our instruction synthesis framework. We design a dialogue skeleton by combining nine types of human conversational intents with corresponding curated information flows. \textit{ConsistentChat} is constructed in two stages: First, we generate a set of multi-turn dialogue queries from Qwen-2.5-72B-Instruct based on the skeleton template; Then, we prompt LLMs to generate the corresponding response set with CoT method. On the right side, we present a synthetic dialogue case, which received a high assessment from Qwen-2.5-72B-Instruct.}
\label{fig:framework}
\end{figure*}

Motivated by the above insights, we present a novel framework for multi-turn instruction generation. First, we provide the background on general dialogue generation (\S\ref{sec:definition}). Then, we introduce the details of Skeleton-Guided Multi-Turn Dialogue Framework (\S\ref{sec:method}).

\subsection{Definition}
\label{sec:definition}

We formalize a multi-turn dialogue instruction as $\mathcal{D}=\{(\mathcal{I}_j,\mathcal{C}_j)\}_{j=1}^{M}$, $M$ represents total number of dialogues. Each dialogue instance consists of: 

(1) An instruction $\mathcal{I}_j$, which provides task-specific context, such as background knowledge, conversational intent, or speaker attributes.

(2) A conversation history $\mathcal{C}_j=\{\langle q_k,r_k\rangle\}_{k=1}^{T_j}$, representing the sequential utterances exchanged between the user query $q_k$ and the conversational model response $r_k$. Here, $T_j$ denotes the total number of dialogue turns in a given interaction.

Given an instruction $\mathcal{I}$ and a dialogue context $\mathcal{C}=\{\langle q_1,r_1\rangle, \langle q_2,r_2\rangle,\cdots,\langle q_t\rangle\}$, the purpose of the dialogue model is to generate an overall model response $r_t$ that aligns with the given context while maintaining chat consistency across turns.

\subsection{Skeleton-Guided Dialogue Framework}
\label{sec:method}
We propose a Skeleton-Guided Multi-Turn Dialogue framework that enhances conversational consistency across multiple turns. An overview of the framework is presented in Figure \ref{fig:framework}, where the left part illustrates the multi-turn instructions synthesis pipeline, and the right part provides a synthetic dialogue case with high consistency.

\begin{figure}[H]
\centering
\includegraphics[width=1\linewidth]{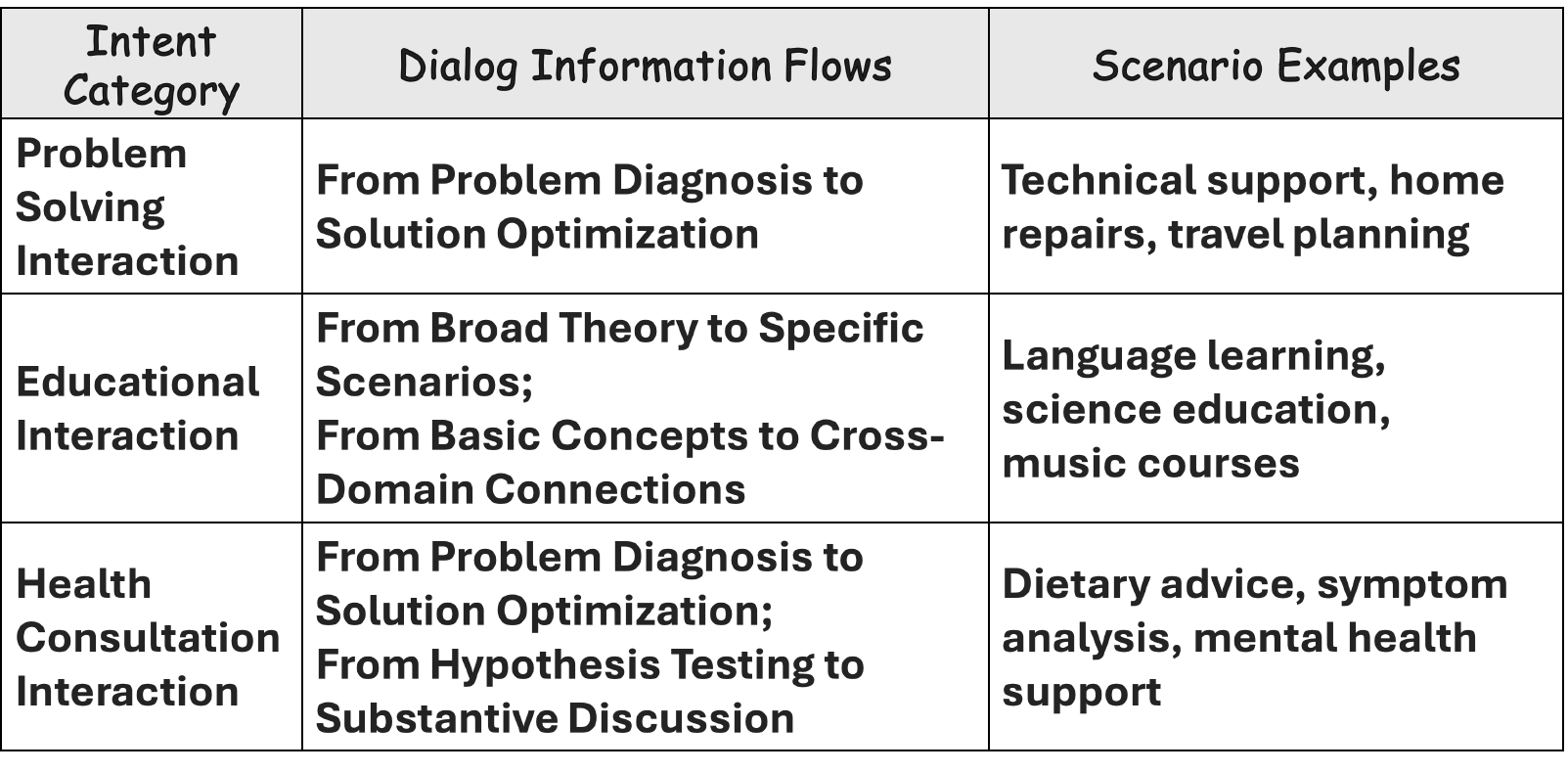}
\caption{Dialogue Skeleton-Guided Schema: examples of interaction intent categories used in our framework. This table presents three categories to illustrate how scenario types align with dialogue flow strategies.}
\label{fig:intent}
\end{figure}

\paragraph{Intent-Based Skeletons}
Our framework stems from an analysis of human conversational patterns in multi-turn interactions. Drawing on theories from human–computer interaction, particularly the Conversational Acts Taxonomy proposed by \citet{rapp2021human}, we identify nine fundamental interaction patterns. These patterns span a wide range of human conversational behaviors and are linked to dynamic information flows that guide the logical progression of dialogue. Further details are provided in Appendix \ref{appendix:framework}, with three examples of conversational intents shown in Figure \ref{fig:intent}.

\paragraph{Query Generation via Human Intent}
\label{sec:query_generation}
We leverage the distillation capability of LLMs through a Skeleton-Guided prompt (see Appendix \ref{appendix:skeleton-guided-prompt}) to generate query sequences that preserve conversational consistency across multiple turns. This design enables the model to capture high-level cues while retaining sufficient flexibility. Unlike traditional multi-turn dialogue generation, our framework ensures that queries progress in a logical order, extending the conversation while avoiding topic drift.

The query generation process follows a structured approach, where each query \( q_t \) is generated sequentially based on a given dialogue intent \( \mathcal{I} \) and its corresponding information flow \( \mathcal{F} \), ensuring that the user interactions adhere to common human dialogue patterns. Formally, we define the probability of generating a sequence of user queries as:  
\begin{equation}
    q_1, q_2, \dots, q_T \sim P(Q \mid \mathcal{I}, \mathcal{F}, \theta_{\text{LLMs}})
\end{equation}  
where \( Q = \{q_1, q_2, ..., q_T\} \) represents the full set of user queries, and \( \theta_{\text{LLMs}} \) denotes the LLMs for distillation. Each query \( q_t \) is generated not only on the intent \( \mathcal{I} \) and information flow \( \mathcal{F} \), but also on previously generated queries to ensure new-generated queries remain aligned with earlier ones, simulating human dialogue evolution. Also, our approach prevents redundant or drifted user inputs, making the synthetic dialogues more consistent.

\paragraph{Response Generation via CoT Prompt}
\label{sec:response_generation}
Once the user queries are generated, we continue to generate all corresponding responses in a single forward pass instead of a turn-by-turn manner. This approach significantly enhances the context-awareness and fluency of the generated dialogue responses. This can be expressed as follows:
\begin{equation}
    r_t \sim P(R \mid Q, r_{<t}, \theta_{\text{LLMs}})
\end{equation}
\( Q \) is the user query set, \( r_{<t} \) represents agent response, and \( R \) denotes the candidate responses. 
Yet, generating responses turn by turn suffers from the limitation in LLMs due to their causal attention mechanism. Since modern LLMs operate under a causal masking constraint, they cannot see future inputs during completion. This means when generating \( r_t \), the model cannot adjust response on future user queries \( q_{>t} \), potentially leading to inconsistencies or unnatural conversational flow.

Thus, we employ single-pass generation strategy, where the model is asked to generate the entire response sequences in one inference step. By using CoT reasoning and full-sequence response generation, multi-turn dialogues are well-planned and contextually aligned, leading to higher quality and more human-like interactions.

\subsection{Instruction Fine-tuning}
Through the two steps above, we build a multi-turn dialogue dataset with  high chat consistency, \textit{ConsistentChat} (details in Appendix \ref{appendix:consistentchat}), which can be used for full-parameter fine-tuning of language models to enhance their consistency in multi-turn conversations. We fine-tune pretrained models on this dataset to improve their performance in maintaining coherence across dialogue turns. As defined by \citet{iyer2022opt}, the loss function is as follows:
\begin{align*}
\small
\mathcal{L}(D_{\boldsymbol{x}}; \theta) = - \sum_{i=1}^{N} \log p_\theta\left(\boldsymbol{y}_i \mid \left[\boldsymbol{x}, \left\{[c_i, d_i]\right\}_{i=1}^{|D_{\boldsymbol{x}}|}\right], \boldsymbol{y}_{<i}\right)
\end{align*}

\section{Experiments}

\subsection{Setup}
\paragraph{Dataset.}
We generated approximately 100 random scenarios for each conversational intent by our proposed pipeline, setting the temperature parameter to 0.9 to increase diversity. We employed open-source Qwen-2.5-72B-Instruct \cite{yang2024qwen2.5} to generate a collection of queries, with each scenario producing more than 15 distinct multi-turn dialogues. Then we generated the corresponding response collection in a single pass. Ultimately, we collected approximately 15,000 multi-turn conversations and 224,392 utterances, characterized by high chat consistency and comprehensive coverage of conversational intents.

\begin{table*}[t!]
\centering
\resizebox{0.95\textwidth}{!}{
\begin{tabular}{l c c c c c}
\toprule
  \multirow{2}[2]{*}{Models} & \multicolumn{2}{c}{\textsc{Light}} & \multicolumn{2}{c}{\textsc{TopDial}} & \multirow{2}[2]{*}{Avg.} \\
\cmidrule(lr){2-3} \cmidrule(lr){4-5}
  & \textsc{Qwen} Score & \textsc{LLaMA} Score & \textsc{Qwen} Score & \textsc{LLaMA} Score & \\
\midrule
Qwen-2.5-72B-Instruct & 7.48 & 7.92 & 7.87 & 8.05 & 7.83 \\
\hdashline
Qwen-2.5-7B & 6.36 & 5.69 & 6.98 & 6.42 & 6.36 \\
Qwen-2.5-7B-ShareGPT & 6.71 & \underline{7.32} & 7.03 & \underline{7.33} & \underline{7.10} \\
Qwen-2.5-7B-ChatAlpaca & 6.11 & 6.97 & 6.70 & 6.87 & 6.66 \\
Qwen-2.5-7B-UltraChat & \underline{6.78} & 7.23 & \underline{7.14} & 6.90 & 7.01 \\
Qwen-2.5-7B-LmsysChat & 6.00 & 6.07 & 6.44 & 5.83 & 6.09 \\
Qwen-2.5-7B-\textit{ConsistentChat} & \textbf{6.94} & \textbf{7.50} & \textbf{7.34} & \textbf{7.51} & \textbf{7.32} \\
\midrule
LLaMA-3.1-70B-Instruct & 7.44 & 7.86 & 7.57 & 7.62 & 7.62 \\
\hdashline
LLaMA-3.1-8B & 4.55 & 3.76 & 5.83 & 5.34 & 4.87 \\
LLaMA-3.1-8B-ShareGPT & \underline{6.42} & \underline{6.66} & 6.62 & 6.39 & 6.52 \\
LLaMA-3.1-8B-ChatAlpaca & 6.38 & 6.56 & 6.85 & 6.77 & 6.64 \\
LLaMA-3.1-8B-UltraChat & 6.15 & 6.55 & \underline{7.14} &\underline{6.84} & \underline{6.67} \\
LLaMA-3.1-8B-LmsysChat & 5.66 & 5.43 & 6.24 & 4.59 & 5.48 \\
LLaMA-3.1-8B-\textit{ConsistentChat} & \textbf{6.71} & \textbf{6.72} & \textbf{7.22} & \textbf{7.06} & \textbf{6.93} \\
\midrule
Mistral-7B-v0.3 & 3.09 & 2.49 & 4.09 & 4.00 & 3.42 \\
Mistral-7B-v0.3-ShareGPT & \underline{6.33} & \textbf{6.71} & 6.71 & 5.61 & \underline{6.34} \\
Mistral-7B-v0.3-ChatAlpaca & 5.65 & 6.18 & 6.22 & 5.20 & 5.81 \\
Mistral-7B-v0.3-UltraChat & 5.49 & 6.08 & \underline{6.83} & \underline{6.36} & 6.19 \\
Mistral-7B-v0.3-LmsysChat & 5.08 & 5.52 & 6.01 & 5.37 & 5.50 \\
Mistral-7B-v0.3-\textit{ConsistentChat} & \textbf{6.62} & \underline{6.21} & \textbf{7.09} & \textbf{6.67} & \textbf{6.65} \\
\bottomrule
\end{tabular}}
\caption{The overall experimental results of \textit{ConsistentChat} and other baselines on the \textsc{Light} and \textsc{TopDial} benchmarks, based on chat consistency metric. \textsc{Qwen} Score and \textsc{LLaMA} Score indicate the results obtained from \textit{Qwen-2.5-72B-Instruct} and \textit{LLaMA-3.1-70B-Instruct}, respectively. The best/second scores are \textbf{bolded}/\underline{underlined}.}
\label{tab:main_exp_result}
\end{table*}

\paragraph{Baselines.}
We compared models fine-tuned on our data with those fine-tuned on several widely-used dialogue datasets. For all baseline datasets, we randomly sampled about 15,000 multi-turn dialogues with more than 3 turns to ensure a fair comparison with \textit{ConsistentChat}. Below, we provide an overview of each baseline dataset:

\textbf{ShareGPT} \cite{vicuna2023}: This dataset is a collection of user-shared conversations with ChatGPT. We utilized the high-quality, cleaned version by \citet{shibing2023sharegpt}, which contains carefully filtered conversations covering a broad range of topics and use cases from GPT-4. 

\textbf{ChatAlpaca} \cite{ChatAlpaca}: It contains 10,000 dialogues generated by prompting GPT-3.5-turbo \cite{openai2022chatgpt} to simulate human-assistant interactions. ChatAlpaca focuses on instruction-following capabilities rather than extended multi-turn consistency.

\textbf{UltraChat} \cite{ding2023enhancing}: Consisting of about 1.5 million conversations, it helps research in building capable and engaging conversational agents. It was created using a taxonomy-guided data generation approach, covering 30 categories and over 40 subcategories of human intentions.

\textbf{Lmsys-Chat-1M (LmsysChat)} \cite{zheng2023lmsyschat1m}: Gathered from interactions with chatbots on the Arena platform, it offered a rich and diverse human-chatbot dialogues across different models and user intents. It contains about 80,000 conversations collected from real-world usage, with diverse query types and interaction patterns.

We performed supervised fine-tuning of the pre-trained models on each dataset to conduct a fair comparison of their ability to enhance chat consistency and multi-turn conversational performance, with particular emphasis on chat consistency and contextual awareness across conversations.

\paragraph{Evaluation.}
We employed three representative multi-turn dialogue benchmarks to evaluate the performance of our models: \textsc{Light} \cite{urbanek-etal-2019-learning} and \textsc{TopDial} \cite{wang-etal-2023-target}, which are used to evaluate chat consistency in multi-turn dialogues, and \textsc{MT-Eval} \cite{kwan2024mt}, which assesses comprehensive multi-turn conversational abilities.

\textsc{\textbf{Light}} is a character-based dialogue dataset, which contains various characters, collected from crowdworker interactions ranging from animals to humans (e.g., snake, seagull, knight). Each dialogue has a background (settings like forest and castle), including persona attributes and characteristics (see detailed examples in Appendix \ref{appendix:light_example}). We utilized the \textit{Test-seen} subset, which comprises 1,000 dialogues and 13,392 utterances.

\textsc{\textbf{TopDial}} is a goal-oriented dataset designed for proactive agents interacting with personalized users. These goals primarily revolve around movies, music, and food. Agents need to actively guide the conversation toward target topics based on domain knowledge and assigned user attributes to maintain engagement rather than abruptly steering the dialogue. The agent acts on the target topic (see detailed examples in Appendix \ref{appendix:topdial_example}). We used the \textit{dialogue\_test\_seen} subset, which contains 3,606 dialogues and 40,496 utterances.

\begin{table}[t!]
\centering
\resizebox{\columnwidth}{!}{
\begin{tabular}{l c c}
\toprule
Models & ST Score & MT Score \\
\midrule
Qwen-2.5-14B-Instruct            & 8.01 & 7.95 (-0.06) \\
\hdashline
Qwen-2.5-7B                      & 5.66 & 5.83 (+0.17) \\
Qwen-2.5-7B-ShareGPT             & 7.81 & 7.86 (+0.05) \\
Qwen-2.5-7B-ChatAlpaca           & \underline{7.86} & \underline{8.12} (+0.26) \\
Qwen-2.5-7B-UltraChat            & 6.18 & 6.65 (+0.47) \\
Qwen-2.5-7B-LmsysChat            & 5.61 & 5.74 (+0.13) \\
Qwen-2.5-7B-\textit{ConsistentChat}
                                 & \textbf{8.07} & \textbf{8.38} (+0.31) \\
\midrule
LLaMA-3.1-8B                     & 4.86 & 4.38 (–0.48) \\
LLaMA-3.1-8B-ShareGPT            & \underline{7.40} & 7.60 (+0.20) \\
LLaMA-3.1-8B-ChatAlpaca          & 7.37 & \underline{7.73} (+0.36) \\
LLaMA-3.1-8B-UltraChat           & 6.89 & 6.85 (–0.04) \\
LLaMA-3.1-8B-LmsysChat           & 5.66 & 5.78 (+0.12) \\
LLaMA-3.1-8B-\textit{ConsistentChat}
                                 & \textbf{7.71} & \textbf{7.93} (+0.22) \\
\midrule
Mistral-7B-v0.3                  & 4.41 & 5.71 (+1.30) \\
Mistral-7B-v0.3-ShareGPT         & 6.39 & \underline{6.94} (+0.55) \\
Mistral-7B-v0.3-ChatAlpaca       & \underline{6.47} & 6.68 (+0.21) \\
Mistral-7B-v0.3-UltraChat        & 5.97 & 6.23 (+0.26) \\
Mistral-7B-v0.3-LmsysChat        & 5.48 & 5.06 (–0.42) \\
Mistral-7B-v0.3-\textit{ConsistentChat}
                                 & \textbf{6.67} & \textbf{7.14} (+0.47) \\
\bottomrule
\end{tabular}
}
\caption{Evaluation results on the \textsc{MT-Eval} benchmark under both single-turn (\textbf{ST}) and multi-turn (\textbf{MT}) conditions. The best/second scores are \textbf{bolded}/\underline{underlined}. Bracketed numbers indicate the change in score between the single-turn and multi-turn scenarios.}
\label{tab:mt-eval}
\end{table}

\textsc{\textbf{MT-Eval}} is designed to evaluate models' multi-turn conversational capabilities, developed through analysis of Human-LLM interactions, and encompasses a wide range of real-world backgrounds. We utilized Qwen-2.5-72B-Instruct models as evaluator and three tasks from this dataset: Expansion, Refinement, and Follow-up, which contain 70, 480 and 240 turns respectively. Expansion involves the exploration of varied topics within the main subject; Refinement focuses on clarifying or revising initial instructions; and Follow-up consists of questions based on the assistant's previous responses.

\paragraph{Implementation Details.}
We utilized three pre-trained language models: Qwen-2.5-7B \cite{yang2024qwen2.5}, LLaMA-3.1-8B \cite{grattafiori2024llama}, and Mistral-7B-v0.3 \cite{Albert2024mistral}, and performed supervised fine-tuning using different datasets. For all models, we implemented a learning rate of 1e-5 with a cosine learning rate schedule for 3 epochs, using per device train batch size of 1 and gradient accumulation steps of 2. We used LLaMA-Factory \cite{zheng2024llamafactory} to perform supervised fine-tuning and employed VLLM \cite{kwon2023efficient} to accelerate inference.

\begin{figure}[H]
\centering
\includegraphics[width=1\linewidth]{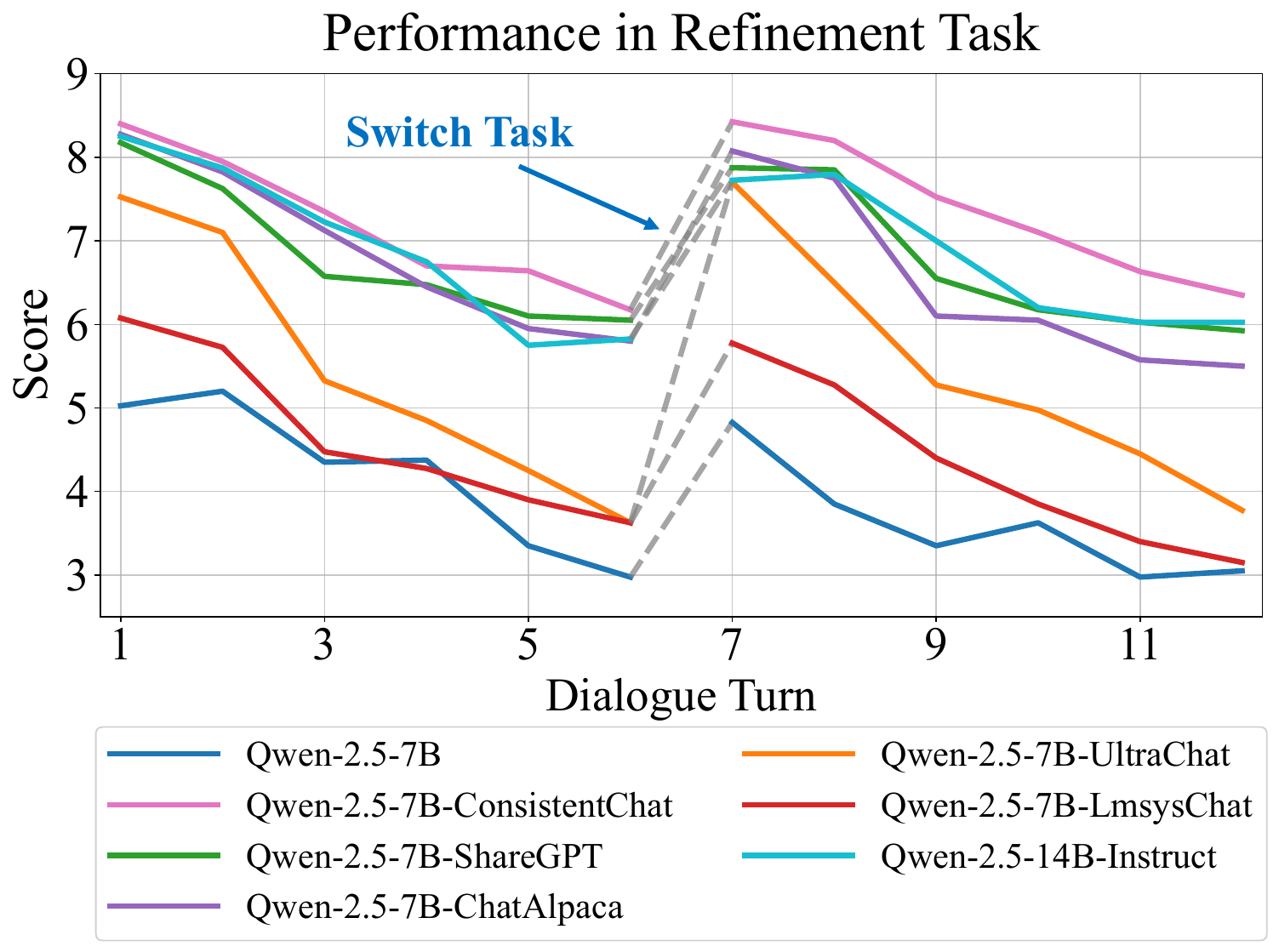}
\caption{Among all the fine-tuned Qwen models evaluated in Refinement task on \textsc{MT-eval}, Qwen-2.5-7B-\textit{ConsistentChat} achieves the state-of-the-art performance, even outperforming Qwen-2.5-14B-Instruct. Each dialogue in task has two NLP tasks with each task comprising six increasingly complex instructions. The transition to the second NLP task occurs at the seventh turn as denoted by the gray dashed line.}
\label{fig:mt-eval}
\end{figure}

\subsection{Overall Results}
Table \ref{tab:main_exp_result} reports the results of chat consistency metrics, while Table \ref{tab:mt-eval} presents multi-turn conversational abilities. Taken together, we can see that:

\begin{table*}[ht]
  \centering
  \resizebox{\textwidth}{!}{
    \begin{tabular}{lcccccccc}
      \toprule
      \multirow{2}[2]{*}{Models} & \multicolumn{3}{c}{Reasoning} & \multicolumn{2}{c}{Examination} & Knowledge & Code & \multirow{2}[2]{*}{\textbf{Avg.}} \\
      \cmidrule(lr){2-4} \cmidrule(lr){5-6} \cmidrule(lr){7-7} \cmidrule(lr){8-8}
       & Hellaswag & MATH & GPQA & MMLU & Gaokao & TriviaQA & HumanEval \\
      \midrule
      Qwen-2.5-7B & 80.20 & 49.80 & \textbf{36.40} & \textbf{74.20} & 61.05 & 64.93 & 57.90 & 60.64 \\
      \hdashline
      Qwen-2.5-7B-ShareGPT & 79.96 & 52.42 & 29.80 & 70.42 & 71.96 & 60.79 & 70.73 & 62.30 \\
      Qwen-2.5-7B-ChatAlpaca & 80.40 & 36.68 & 32.32 & 73.66 & 74.13 & 66.49 & 67.68 & 61.62\\
      Qwen-2.5-7B-UltraChat & 70.56 & 46.26 & 32.83 & 71.36 & 71.88 & 60.94 & 58.54 & 58.91 \\
      Qwen-2.5-7B-LmsysChat & 48.19 & 14.40 & 30.81 & 59.48 & 30.66 & 46.29 & 32.32 & 37.44 \\
      Qwen-2.5-7B-\textit{ConsistentChat} & \textbf{83.09} & \textbf{64.96} & 35.35 & 74.02 & \textbf{75.54} & \textbf{67.31} & \textbf{77.44} & \textbf{68.24}\\
      \bottomrule
    \end{tabular}
  }
  \caption{Results of fine-tuning Qwen models on commonsense reasoning tasks. Qwen-2.5-7B-\textit{ConsistentChat} exhibits stronger general capabilities, with notable improvements in MATH($\uparrow$ \textbf{30.4\%}) and HumanEval($\uparrow$ \textbf{33.7\%}) compared to the pre-trained model, indicating an average gain of \textbf{12.5\%} across all commonsense benchmarks.}
  \label{tab:commonsense_results}
\end{table*}

\paragraph{Superior Chat Consistency with ConsistentChat}
Fine-tuning with \textit{ConsistentChat} achieves the highest average scores on chat consistency metrics across both the \textsc{Light} and \textsc{TopDial} in Table \ref{tab:main_exp_result}. This demonstrates that our framework effectively generates high-quality instructions that helps models maintain coherent topics throughout conversations, outperforming other widely-used datasets.

\paragraph{Enhanced Multi-turn Conversational Ability}
Models trained with \textit{ConsistentChat} also demonstrate improved multi-turn conversational capabilities. As shown in Table \ref{tab:mt-eval}, these models outperform all the models trained on baseline datasets in both single-turn (ST) and multi-turn (MT) conditions, even outperforming Qwen-2.5-14B-Instruct. This indicates that fine-tuning with \textit{ConsistentChat} yields models that are not only more consistent but also better at managing extended conversations, validating our strategy of comprehensive intent coverage and consistency during data synthesis.

\paragraph{Positive Correlation between Chat Consistency and Multi-turn Performance}
The strong performance of models trained with \textit{ConsistentChat} in both chat consistency and multi-turn dialogue suggests a positive correlation: Enhancing consistency substantially improves overall multi-turn competence, underscoring its importance for building robust conversational AI systems, such as intelligent customer service and task-oriented dialogue agents.

\subsection{Human Evaluation}
To verify the automatic evaluation aligns with human preferences, we conducted an evaluation following recent studies \cite{zheng2025pptagent,kwan2024mt}. We recruited three well-educated graduate students as annotators to score 50 randomly selected multi-turn dialogues from each benchmark. Further details are provided in Appendix \ref{appendix:human_evaluation}

As shown in Table \ref{tab:human_eval}, we report that both Pearson's correlation and Spearman's rank correlation between human annotators and LLM. The average Spearman correlation reaches 0.68, surpassing alternative evaluation methods for natural language generation tasks \cite{liu2023g}. These results suggest the LLM-as-a-Judge method provides ratings well aligned with human, consistent with recent findings \cite{kwan2024mt}.

\subsection{Commonsense Results}
To further validate our method, we used Qwen models to compare general abilities in commonsense reasoning across the reasoning, examination, knowledge, and code domains. Table \ref{tab:commonsense_results} illustrates that Qwen-2.5-7B-\textit{ConsistentChat} achieves better or competitive performance across 7 commonsense reasoning datasets, exhibiting great improvements in MATH and Coding capabilities (49.80 vs. 64.96, $\uparrow$ \textbf{30.4\%}; 57.90 vs. 77.44, $\uparrow$ \textbf{33.7\%}), while other models show minimal improvements or even degradation compared to their base models. These improvements can be attributed to our thorough analysis of human dialogue intents and the diversity of training instructions.

\begin{table}[ht]
  \centering
  \resizebox{0.68\columnwidth}{!}{ 
  \begin{tabular}{lcc}
    \toprule
    Task & Pearson & Spearman  \\
    \midrule
    \textsc{Light} & 0.73 & 0.70  \\
    \textsc{TopDial} & 0.69 & 0.66  \\
    \midrule
    \textbf{Avg.} & 0.71 & 0.68  \\
    \bottomrule
  \end{tabular}}
  \caption{The correlation scores between human ratings and Qwen-2.5-72B-Instruct ratings for \textsc{Light} and \textsc{TopDial} for consistency. The results demonstrate strong alignment between automatic and human evaluations.}
  \label{tab:human_eval}
\end{table}

\subsection{Quality Analysis of Generated Data}
This section analyzes the quality of automatically generated instructions along two dimensions: the overall diversity and similarity in dialogues.

\paragraph{Diversity of Generated Instruction}
Figure \ref{fig:diverse_ours} in Appendix presents a sunburst visualization of verb–noun structures, following the methodology of Self-Instruct \cite{wang-etal-2023-self-instruct} to illustrate the diversity of instructions. Our results show that the generated instructions cover all major human interaction types commonly expected from AI assistants, including information seeking, planning, problem solving and creative ideation. This balanced, hierarchical distribution demonstrates that our approach effectively captures a wide range of user–assistant conversational needs.

\begin{table}[htbp]
\centering
\resizebox{0.98\columnwidth}{!}{ 
\begin{tabular}{lccc}
\toprule
Dataset & DistilBERT & MPNet & MiniLM \\
\midrule
ShareGPT   & 0.892  & 0.895 & 0.857 \\
ChatAlpaca & 0.851  & 0.874 & 0.862  \\
UltraChat  & 0.907  & 0.911 & 0.915  \\
LmsysChat  & 0.817  & 0.847 & 0.813 \\
\textbf{\textit{ConsistentChat}}  & \textbf{0.915}  & \textbf{0.919 }& \textbf{0.923} \\
\bottomrule
\end{tabular}}
\caption{\textit{ConsistentChat} dataset surpasses other datasets across all embedding models, demonstrating stronger alignment and semantic consistency in the user utterances of multi-turn dialogues.}
\label{tab:embedding}
\end{table}
\paragraph{Similarity of Generated Instruction in dialogue}
To evaluate the consistency of SFT datasets, we computed the average semantic similarity between user queries in each conversation. Using sentence embedding models including DistilBERT (\href{https://huggingface.co/sentence-transformers/multi-qa-distilbert-cos-v1}{\texttt{multi-qa-distilbert-cos-v1}}), MPNet (\href{https://huggingface.co/sentence-transformers/multi-qa-mpnet-base-dot-v1}{\texttt{multi-qa-mpnet-base-dot-v1}}), and MiniLM (\href{https://huggingface.co/sentence-transformers/multi-qa-MiniLM-L6-cos-v1}{\texttt{multi-qa-MiniLM-L6-cos-v1}}), we incorporated user queries to capture intent continuity. As shown in Table \ref{tab:embedding}, \textit{ConsistentChat} surpasses across all embedding models, demonstrating stronger alignment and semantic consistency in user queries.

\section{Related Work}

\paragraph{Dialogue Language Models}
Recent advancements in LLMs have revolutionized conversational AI through large-scale pre-training on diverse fine-tuning instructions \cite{chen2022wavlm, brown2020language}. A major breakthrough was achieved with the dawn of ChatGPT \cite{openai2022chatgpt}, which incorporated instruction tuning and alignment techniques to improve response quality and contextual awareness. Dialogue models such as LLaMA-3.1 \cite{grattafiori2024llama}, Qwen-2.5 \cite{yang2024qwen2.5} demonstrate emerging dialogue capabilities through instruction tuning on single-turn or multi-turn interaction. These models typically rely on auto-regressive generation, where responses at each turn are conditioned on prior context. Tuning language models has become a prevalent paradigm for building capable dialogue agents \cite{iyer2022opt}. Meanwhile, continual learning techniques such as Self-Synthesized Rehearsal \cite{huang2024mitigating} address catastrophic forgetting in LLMs, enabling more stable model adaptation across evolving dialogue tasks. These work mainly focus on fine-tuning open-source LLMs for dialogue generation.

\paragraph{Chat Consistency in Dialogue}
The chat consistency of a dialogue refers to the ability of a conversational agent to generate responses that remain consistent with the dialogue history, its assigned role \cite{urbanek-etal-2019-learning, shuster-etal-2022-state,chen-etal-2023-large}, and the overall discussion topic \cite{wang-etal-2017-steering, wang-etal-2024-instruct}. Preserving chat consistency remains an open challenge in multi-turn dialogue modeling. Recent works such as FaithfulRAG \cite{zhang2025faithfulrag} explicitly model fact-level conflicts between parametric knowledge and retrieved context, thereby motivating methods to reason about and resolve inconsistencies prior to response generation. However, existing multiple dialogue datasets inevitably own topic drifts from human conversations, unconsciously training models to replicate inconsistent behaviors. \citet{wang-etal-2024-instruct} proposes \textsc{Midi}-Tuning, separately modeling the user and agent to improve chat consistency remarkably. Also, latest works such as \citet{zhang2025notall}, \citet{han2025attributes} and \citet{shi2025personax} focus on enhancing the quality of synthetic data and have achieved promising results. Our work bridges this gap by introducing Skeleton-Guided instruction generation method, a novel paradigm combining human conversational intents with structured information flow constraints to improve the consistency in multi-turn dialogues remarkably.

\section{Conclusion}
In this paper, we propose \textit{ConsistentChat} dataset, and an instruction synthesis framework to automatically generate multi-turn dialogues with high chat consistency. Our method addresses a wide spectrum of human conversational intents without reliance on proprietary LLMs. Experimental results show that models fine-tuned on \textit{ConsistentChat} exhibit superior chat consistency in dialogues and comprehensive multi-turn conversational capabilities, validating the quality of our synthetic instructions. We believe this work provides insights for instruction synthesis and helps foster research in instruction tuning. For future work, we aim to conduct a more in-depth study of human conversational intents to achieve even broader coverage and further enhance the quality of the instructions.

\section*{Limitations}
There are several limitations of our current framework, which we plan to address in the future. Firstly, while LLMs-based evaluations align with human ratings, residual error and internal bias highlight the need for comprehensive human validation. Secondly, due to computational constraints, the applicability of our framework to larger models remains to be explored. Thirdly, despite capability gains, further alignment is required to advance the models' capabilities.

\section*{Acknowledgements}
We sincerely thank the reviewers for their insightful comments and valuable suggestions. This work was supported by Beijing Natural Science Foundation (L243006), the Natural Science Foundation of China (No. 62272439, 62306303, 62476265).

\bibliography{custom}

\begin{thebibliography}{37}
\providecommand{\natexlab}[1]{#1}

\bibitem[{Bian et~al.(2023)Bian, Lin, Lu, Han, Sun, and He}]{ChatAlpaca}
Ning Bian, Hongyu Lin, Yaojie Lu, Xianpei Han, Le~Sun, and Ben He. 2023.
\newblock Chatalpaca: A multi-turn dialogue corpus based on alpaca
  instructions.
\newblock \url{https://github.com/cascip/ChatAlpaca}.

\bibitem[{Brown et~al.(2020)Brown, Mann, Ryder, Subbiah, Kaplan, Dhariwal,
  Neelakantan, Shyam, Sastry, Askell et~al.}]{brown2020language}
Tom Brown, Benjamin Mann, Nick Ryder, Melanie Subbiah, Jared~D Kaplan, Prafulla
  Dhariwal, Arvind Neelakantan, Pranav Shyam, Girish Sastry, Amanda Askell,
  et~al. 2020.
\newblock Language models are few-shot learners.
\newblock \emph{Advances in Neural Information Processing Systems},
  33:1877--1901.

\bibitem[{Chen et~al.(2023)Chen, Wang, Jiang, Cai, Li, Chen, Wang, and
  Li}]{chen-etal-2023-large}
Nuo Chen, Yan Wang, Haiyun Jiang, Deng Cai, Yuhan Li, Ziyang Chen, Longyue
  Wang, and Jia Li. 2023.
\newblock \href {https://doi.org/10.18653/v1/2023.findings-emnlp.570} {Large
  language models meet harry potter: A dataset for aligning dialogue agents
  with characters}.
\newblock In \emph{Findings of the Association for Computational Linguistics:
  EMNLP 2023}, pages 8506--8520, Singapore. Association for Computational
  Linguistics.

\bibitem[{Chen et~al.(2022)Chen, Wang, Chen, Wu, Liu, Chen, Li, Kanda,
  Yoshioka, Xiao et~al.}]{chen2022wavlm}
Sanyuan Chen, Chengyi Wang, Zhengyang Chen, Yu~Wu, Shujie Liu, Zhuo Chen, Jinyu
  Li, Naoyuki Kanda, Takuya Yoshioka, Xiong Xiao, et~al. 2022.
\newblock Wavlm: Large-scale self-supervised pre-training for full stack speech
  processing.
\newblock \emph{IEEE Journal of Selected Topics in Signal Processing},
  16(6):1505--1518.

\bibitem[{Chen et~al.(2024)Chen, Guan, Lu, Lin, Han, and
  Sun}]{chen2024reinstruct}
Shu Chen, Xinyan Guan, Yaojie Lu, Hongyu Lin, Xianpei Han, and Le~Sun. 2024.
\newblock Reinstruct: Building instruction data from unlabeled corpus.
\newblock In \emph{Findings of the Association for Computational Linguistics
  ACL 2024}, pages 6840--6856.

\bibitem[{Chiang et~al.(2023)Chiang, Li, Lin, Sheng, Wu, Zhang, Zheng, Zhuang,
  Zhuang, Gonzalez, Stoica, and Xing}]{vicuna2023}
Wei-Lin Chiang, Zhuohan Li, Zi~Lin, Ying Sheng, Zhanghao Wu, Hao Zhang, Lianmin
  Zheng, Siyuan Zhuang, Yonghao Zhuang, Joseph~E. Gonzalez, Ion Stoica, and
  Eric~P. Xing. 2023.
\newblock \href {https://lmsys.org/blog/2023-03-30-vicuna/} {Vicuna: An
  open-source chatbot impressing gpt-4 with 90\%* chatgpt quality}.

\bibitem[{Ding et~al.(2023)Ding, Chen, Xu, Qin, Zheng, Hu, Liu, Sun, and
  Zhou}]{ding2023enhancing}
Ning Ding, Yulin Chen, Bokai Xu, Yujia Qin, Zhi Zheng, Shengding Hu, Zhiyuan
  Liu, Maosong Sun, and Bowen Zhou. 2023.
\newblock Enhancing chat language models by scaling high-quality instructional
  conversations.
\newblock \emph{arXiv preprint arXiv:2305.14233}.

\bibitem[{Grattafiori et~al.(2024)Grattafiori, Dubey, Jauhri, Pandey, Kadian,
  Al-Dahle, Letman, Mathur, Schelten, Vaughan et~al.}]{grattafiori2024llama}
Aaron Grattafiori, Abhimanyu Dubey, Abhinav Jauhri, Abhinav Pandey, Abhishek
  Kadian, Ahmad Al-Dahle, Aiesha Letman, Akhil Mathur, Alan Schelten, Alex
  Vaughan, et~al. 2024.
\newblock The llama 3 herd of models.
\newblock \emph{arXiv preprint arXiv:2407.21783}.

\bibitem[{Han et~al.(2025)Han, Liu, and Huang}]{han2025attributes}
Guangzeng Han, Weisi Liu, and Xiaolei Huang. 2025.
\newblock \href {https://arxiv.org/abs/2509.02040} {Attributes as textual
  genes: Leveraging llms as genetic algorithm simulators for conditional
  synthetic data generation}.
\newblock \emph{Preprint}, arXiv:2509.02040.

\bibitem[{Huang et~al.(2024)Huang, Cui, Wang, Yang, Liao, Song, Yao, and
  Su}]{huang2024mitigating}
Jianheng Huang, Leyang Cui, Ante Wang, Chengyi Yang, Xinting Liao, Linfeng
  Song, Junfeng Yao, and Jinsong Su. 2024.
\newblock Mitigating catastrophic forgetting in large language models with
  self-synthesized rehearsal.
\newblock In \emph{Proceedings of the 62nd Annual Meeting of the Association
  for Computational Linguistics (Volume 1: Long Papers)}, pages 1416--1428.

\bibitem[{Iyer et~al.(2022)Iyer, Lin, Pasunuru, Mihaylov, Simig, Yu, Shuster,
  Wang, Liu, Koura et~al.}]{iyer2022opt}
Srinivasan Iyer, Xi~Victoria Lin, Ramakanth Pasunuru, Todor Mihaylov, Daniel
  Simig, Ping Yu, Kurt Shuster, Tianlu Wang, Qing Liu, Punit~Singh Koura,
  et~al. 2022.
\newblock Opt-iml: Scaling language model instruction meta learning through the
  lens of generalization.
\newblock \emph{arXiv preprint arXiv:2212.12017}.

\bibitem[{Jiang et~al.(2024)Jiang, Sablayrolles, Tacnet, Roux, Mensch,
  Herblin-Stoop, Bout et~al.}]{Albert2024mistral}
Albert Jiang, Alexandre Sablayrolles, Alexis Tacnet, Antoine Roux, Arthur
  Mensch, Audrey Herblin-Stoop, Baptiste Bout, et~al. 2024.
\newblock \href {https://huggingface.co/mistralai/Mistral-7B-v0.3}
  {Mistralai/mistral-7b-v0.3}.

\bibitem[{Ju et~al.(2022)Ju, Feng, Lv, Wang, and Zhang}]{ju2022learning}
Dongshi Ju, Shi Feng, Pengcheng Lv, Daling Wang, and Yifei Zhang. 2022.
\newblock Learning to improve persona consistency in multi-party dialogue
  generation via text knowledge enhancement.
\newblock In \emph{Proceedings of the 29th international conference on
  computational linguistics}, pages 298--309.

\bibitem[{Kwan et~al.(2024)Kwan, Zeng, Jiang, Wang, Li, Shang, Jiang, Liu, and
  Wong}]{kwan2024mt}
Wai-Chung Kwan, Xingshan Zeng, Yuxin Jiang, Yufei Wang, Liangyou Li, Lifeng
  Shang, Xin Jiang, Qun Liu, and Kam-Fai Wong. 2024.
\newblock Mt-eval: A multi-turn capabilities evaluation benchmark for large
  language models.
\newblock In \emph{Proceedings of the 2024 Conference on Empirical Methods in
  Natural Language Processing}, pages 20153--20177.

\bibitem[{Kwon et~al.(2023)Kwon, Li, Zhuang, Sheng, Zheng, Yu, Gonzalez, Zhang,
  and Stoica}]{kwon2023efficient}
Woosuk Kwon, Zhuohan Li, Siyuan Zhuang, Ying Sheng, Lianmin Zheng, Cody~Hao Yu,
  Joseph Gonzalez, Hao Zhang, and Ion Stoica. 2023.
\newblock Efficient memory management for large language model serving with
  pagedattention.
\newblock In \emph{Proceedings of the 29th Symposium on Operating Systems
  Principles}, pages 611--626.

\bibitem[{Li et~al.(2024)Li, Dong, Tang, Wang, Zhang, Huang, Huang, Huang,
  Huang, Zhang et~al.}]{li2024synthetic}
Haoran Li, Qingxiu Dong, Zhengyang Tang, Chaojun Wang, Xingxing Zhang, Haoyang
  Huang, Shaohan Huang, Xiaolong Huang, Zeqiang Huang, Dongdong Zhang, et~al.
  2024.
\newblock Synthetic data (almost) from scratch: Generalized instruction tuning
  for language models.
\newblock \emph{arXiv preprint arXiv:2402.13064}.

\bibitem[{Liu et~al.(2023)Liu, Iter, Xu, Wang, Xu, and Zhu}]{liu2023g}
Yang Liu, Dan Iter, Yichong Xu, Shuohang Wang, Ruochen Xu, and Chenguang Zhu.
  2023.
\newblock G-eval: Nlg evaluation using gpt-4 with better human alignment.
\newblock \emph{arXiv preprint arXiv:2303.16634}.

\bibitem[{Lu et~al.(2023)Lu, An, Lin, Pergola, He, Yin, Sun, and
  Wu}]{lu2023memochat}
Junru Lu, Siyu An, Mingbao Lin, Gabriele Pergola, Yulan He, Di~Yin, Xing Sun,
  and Yunsheng Wu. 2023.
\newblock Memochat: Tuning llms to use memos for consistent long-range
  open-domain conversation.
\newblock \emph{arXiv preprint arXiv:2308.08239}.

\bibitem[{OpenAI(2022)}]{openai2022chatgpt}
OpenAI. 2022.
\newblock Introducing {ChatGPT}.
\newblock \url{https://openai.com/blog/chatgpt}.

\bibitem[{Rapp et~al.(2021)Rapp, Curti, and Boldi}]{rapp2021human}
Amon Rapp, Lorenzo Curti, and Arianna Boldi. 2021.
\newblock The human side of human-chatbot interaction: A systematic literature
  review of ten years of research on text-based chatbots.
\newblock \emph{International Journal of Human-Computer Studies}, 151:102630.

\bibitem[{Sanh et~al.(2022)Sanh, Webson, Raffel, Bach, Sutawika, Alyafeai,
  Chaffin, Stiegler, Le~Scao, Raja et~al.}]{sanh2022multitask}
Victor Sanh, Albert Webson, Colin Raffel, Stephen~H Bach, Lintang Sutawika,
  Zaid Alyafeai, Antoine Chaffin, Arnaud Stiegler, Teven Le~Scao, Arun Raja,
  et~al. 2022.
\newblock Multitask prompted training enables zero-shot task generalization.
\newblock In \emph{ICLR 2022-Tenth International Conference on Learning
  Representations}.

\bibitem[{Shi et~al.(2025)Shi, Xu, Zhang, Zi, Wu, and Xu}]{shi2025personax}
Yunxiao Shi, Wujiang Xu, Zeqi Zhang, Xing Zi, Qiang Wu, and Min Xu. 2025.
\newblock Personax: A recommendation agent oriented user modeling framework for
  long behavior sequence.
\newblock \emph{arXiv preprint arXiv:2503.02398}.

\bibitem[{Shuster et~al.(2022)Shuster, Urbanek, Szlam, and
  Weston}]{shuster-etal-2022-state}
Kurt Shuster, Jack Urbanek, Arthur Szlam, and Jason Weston. 2022.
\newblock \href {https://doi.org/10.18653/v1/2022.findings-naacl.182} {Am {I}
  me or you? state-of-the-art dialogue models cannot maintain an identity}.
\newblock In \emph{Findings of the Association for Computational Linguistics:
  NAACL 2022}, pages 2367--2387, Seattle, United States. Association for
  Computational Linguistics.

\bibitem[{Urbanek et~al.(2019)Urbanek, Fan, Karamcheti, Jain, Humeau, Dinan,
  Rockt{\"a}schel, Kiela, Szlam, and Weston}]{urbanek-etal-2019-learning}
Jack Urbanek, Angela Fan, Siddharth Karamcheti, Saachi Jain, Samuel Humeau,
  Emily Dinan, Tim Rockt{\"a}schel, Douwe Kiela, Arthur Szlam, and Jason
  Weston. 2019.
\newblock \href {https://doi.org/10.18653/v1/D19-1062} {Learning to speak and
  act in a fantasy text adventure game}.
\newblock In \emph{Proceedings of the 2019 Conference on Empirical Methods in
  Natural Language Processing and the 9th International Joint Conference on
  Natural Language Processing (EMNLP-IJCNLP)}, pages 673--683, Hong Kong,
  China. Association for Computational Linguistics.

\bibitem[{Wang et~al.(2017)Wang, Jojic, Brockett, and
  Nyberg}]{wang-etal-2017-steering}
Di~Wang, Nebojsa Jojic, Chris Brockett, and Eric Nyberg. 2017.
\newblock \href {https://doi.org/10.18653/v1/D17-1228} {Steering output style
  and topic in neural response generation}.
\newblock In \emph{Proceedings of the 2017 Conference on Empirical Methods in
  Natural Language Processing}, pages 2140--2150, Copenhagen, Denmark.
  Association for Computational Linguistics.

\bibitem[{Wang et~al.(2023{\natexlab{a}})Wang, Cheng, Lin, Leong, and
  Li}]{wang-etal-2023-target}
Jian Wang, Yi~Cheng, Dongding Lin, Chak Leong, and Wenjie Li.
  2023{\natexlab{a}}.
\newblock \href {https://doi.org/10.18653/v1/2023.emnlp-main.72}
  {Target-oriented proactive dialogue systems with personalization: Problem
  formulation and dataset curation}.
\newblock In \emph{Proceedings of the 2023 Conference on Empirical Methods in
  Natural Language Processing}, pages 1132--1143, Singapore. Association for
  Computational Linguistics.

\bibitem[{Wang et~al.(2024)Wang, Leong, Wang, Lin, Li, and
  Wei}]{wang-etal-2024-instruct}
Jian Wang, Chak~Tou Leong, Jiashuo Wang, Dongding Lin, Wenjie Li, and Xiao-Yong
  Wei. 2024.
\newblock Instruct once, chat consistently in multiple rounds: An efficient
  tuning framework for dialogue.
\newblock In \emph{Proceedings of the 62nd Annual Meeting of the Association
  for Computational Linguistics (ACL)}.

\bibitem[{Wang et~al.(2023{\natexlab{b}})Wang, Kordi, Mishra, Liu, Smith,
  Khashabi, and Hajishirzi}]{wang2023self}
Yizhong Wang, Yeganeh Kordi, Swaroop Mishra, Alisa Liu, Noah~A Smith, Daniel
  Khashabi, and Hannaneh Hajishirzi. 2023{\natexlab{b}}.
\newblock Self-instruct: Aligning language models with self-generated
  instructions.
\newblock In \emph{Proceedings of the 61st Annual Meeting of the Association
  for Computational Linguistics (Volume 1: Long Papers)}, pages 13484--13508.

\bibitem[{Wang et~al.(2023{\natexlab{c}})Wang, Kordi, Mishra, Liu, Smith,
  Khashabi, and Hajishirzi}]{wang-etal-2023-self-instruct}
Yizhong Wang, Yeganeh Kordi, Swaroop Mishra, Alisa Liu, Noah~A. Smith, Daniel
  Khashabi, and Hannaneh Hajishirzi. 2023{\natexlab{c}}.
\newblock \href {https://doi.org/10.18653/v1/2023.acl-long.754} {Self-instruct:
  Aligning language models with self-generated instructions}.
\newblock In \emph{Proceedings of the 61st Annual Meeting of the Association
  for Computational Linguistics (Volume 1: Long Papers)}, pages 13484--13508,
  Toronto, Canada. Association for Computational Linguistics.

\bibitem[{Xu et~al.(2023)Xu, Sun, Zheng, Geng, Zhao, Feng, Tao, and
  Jiang}]{xu2023wizardlm}
Can Xu, Qingfeng Sun, Kai Zheng, Xiubo Geng, Pu~Zhao, Jiazhan Feng, Chongyang
  Tao, and Daxin Jiang. 2023.
\newblock Wizardlm: Empowering large language models to follow complex
  instructions.
\newblock \emph{arXiv preprint arXiv:2304.12244}.

\bibitem[{Xu(2023)}]{shibing2023sharegpt}
Ming Xu. 2023.
\newblock Sharegpt\_gpt4\_version\_dataset.
\newblock \url{https://huggingface.co/datasets/shibing624/sharegpt_gpt4}.

\bibitem[{Yang et~al.(2024)Yang, Yang, Zhang, Hui, Zheng, Yu, Li, Liu, Huang,
  Wei et~al.}]{yang2024qwen2.5}
An~Yang, Baosong Yang, Beichen Zhang, Binyuan Hui, Bo~Zheng, Bowen Yu,
  Chengyuan Li, Dayiheng Liu, Fei Huang, Haoran Wei, et~al. 2024.
\newblock Qwen2.5 technical report.
\newblock \emph{arXiv preprint arXiv:2412.15115}.

\bibitem[{Zhang et~al.(2025{\natexlab{a}})Zhang, Xiang, Xiao, Wang, Li, Wang,
  and Su}]{zhang2025faithfulrag}
Qinggang Zhang, Zhishang Xiang, Yilin Xiao, Le~Wang, Junhui Li, Xinrun Wang,
  and Jinsong Su. 2025{\natexlab{a}}.
\newblock Faithfulrag: Fact-level conflict modeling for context-faithful
  retrieval-augmented generation.
\newblock \emph{arXiv preprint arXiv:2506.08938}.

\bibitem[{Zhang et~al.(2025{\natexlab{b}})Zhang, Yang, Prenkaj, and
  Kasneci}]{zhang2025notall}
Zheyu Zhang, Shuo Yang, Bardh Prenkaj, and Gjergji Kasneci. 2025{\natexlab{b}}.
\newblock Not all features deserve attention: Graph-guided dependency learning
  for tabular data generation with language models.
\newblock \emph{arXiv preprint arXiv:2507.18504}.

\bibitem[{Zheng et~al.(2025)Zheng, Guan, Kong, Zheng, Zhou, Lin, Lu, He, Han,
  and Sun}]{zheng2025pptagent}
Hao Zheng, Xinyan Guan, Hao Kong, Jia Zheng, Weixiang Zhou, Hongyu Lin, Yaojie
  Lu, Ben He, Xianpei Han, and Le~Sun. 2025.
\newblock \href {https://arxiv.org/abs/2501.03936} {Pptagent: Generating and
  evaluating presentations beyond text-to-slides}.
\newblock \emph{Preprint}, arXiv:2501.03936.

\bibitem[{Zheng et~al.(2023)Zheng, Chiang, Sheng, Li, Zhuang, Wu, Zhuang, Li,
  Lin, Xing, Gonzalez, Stoica, and Zhang}]{zheng2023lmsyschat1m}
Lianmin Zheng, Wei-Lin Chiang, Ying Sheng, Tianle Li, Siyuan Zhuang, Zhanghao
  Wu, Yonghao Zhuang, Zhuohan Li, Zi~Lin, Eric.~P Xing, Joseph~E. Gonzalez, Ion
  Stoica, and Hao Zhang. 2023.
\newblock \href {https://arxiv.org/abs/2309.11998} {Lmsys-chat-1m: A
  large-scale real-world llm conversation dataset}.
\newblock \emph{Preprint}, arXiv:2309.11998.

\bibitem[{Zheng et~al.(2024)Zheng, Zhang, Zhang, Ye, Luo, Feng, and
  Ma}]{zheng2024llamafactory}
Yaowei Zheng, Richong Zhang, Junhao Zhang, Yanhan Ye, Zheyan Luo, Zhangchi
  Feng, and Yongqiang Ma. 2024.
\newblock \href {http://arxiv.org/abs/2403.13372} {Llamafactory: Unified
  efficient fine-tuning of 100+ language models}.
\newblock In \emph{Proceedings of the 62nd Annual Meeting of the Association
  for Computational Linguistics (Volume 3: System Demonstrations)}, Bangkok,
  Thailand. Association for Computational Linguistics.

\end{thebibliography}

\newtcolorbox{AIBox}[2][]{aibox,title=#2,#1}

\tcbset{
  aibox/.style={
    width=0.95\textwidth,
    top=5pt,
    colback=black!05,
    colframe=black!20,
    colbacktitle=black!50,
    enhanced,
    center,
    attach boxed title to top left={yshift=-0.1in,xshift=0.1in},
    boxed title style={boxrule=0pt,colframe=white,},
  }
}
\clearpage
\appendix
\label{sec:appendix}
\onecolumn
\section{Experimental Details}
\label{appendix:preliminary}

\subsection{Preliminary Analysis I}
\label{appendix:preliminary_analysis_1}
\paragraph{Performance in \textsc{Light} and \textsc{TopDial}}
We evaluated several popular chat language models (range from 7B to 72B) on \textsc{Light} and \textsc{TopDial} for chat consistency (Table \ref{tab:pre_exp_1}). Most of the 7B chat models achieve average scores between 6 and 7 points (with a full score of 10), suggesting substantial improvement remains. Since Qwen-2.5-72B-Instruct achieves the state-of-the-art performance, we prefer to use it for judging the multi-turn conversational capability on \textsc{MT-Eval}.

\begin{table}[H]
\centering
\resizebox{0.95\columnwidth}{!}{
\begin{tabular}{l c c c c c}
\toprule
  \multirow{2}[2]{*}{Models} & \multicolumn{2}{c}{\textsc{Light}} & \multicolumn{2}{c}{\textsc{TopDial}} & \multirow{2}[2]{*}{\textbf{Avg.}} \\
\cmidrule(lr){2-3} \cmidrule(lr){4-5}
  & \textsc{Qwen} Score & \textsc{LLaMA} Score & \textsc{Qwen} Score & \textsc{LLaMA} Score & \\
\midrule
Qwen-2.5-72B-Instruct & 7.53 & 8.16 & 7.87 & 8.05 & 7.90 \\
LLaMA-3.1-70B-Instruct & 7.44 & 7.86 & 7.57 & 7.62 & 7.62 \\
\hdashline
Gemma-7B-it & 6.16 & 5.57 & 7.23 & 5.71 & 6.17 \\
Vicuna-7B-v1.3 & 6.37 & 6.81 & 6.66 & 5.70 & 6.39 \\
Vicuna-7B-v1.5 & 6.48 & 7.03 & 7.03 & 6.65 & 6.80 \\
Baichuan-2-7B-Chat & 6.69 & 7.23 & 7.05 & 6.68 & 6.91 \\
Qwen-2.5-7B-Instruct & 7.28 & 7.92 & 7.59 & 7.71 & 7.63 \\
\bottomrule
\end{tabular}}
\caption{Results of popular chat models on the \textsc{Light} and \textsc{TopDial} benchmarks, using chat consistency metric. \textsc{Qwen} Score and \textsc{LLaMA} Score indicate the results obtained from \textit{Qwen-2.5-72B-Instruct} and \textit{LLaMA-3.1-70B-Instruct}, respectively. Qwen-2.5-72B-Instruct achieves the state-of-the-art performance.}
\label{tab:pre_exp_1}
\end{table}

\paragraph{Performance in Refinement task}
\label{appendix:pre_mt_eval}
Building on these insights, we evaluate recent open-source models, including Qwen-2.5-Instruct and LLaMA-3.1-Instruct (ranging from 7B to 72B parameters), using the Refinement task from the \textsc{MT-Eval} benchmark. Each dialogue in this task consists of two NLP tasks, each with six increasingly complex instructions. The Refinement task effectively measures how well models follow instructions as dialogue turns progress. Table \ref{tab:pre_exp_4} indicates scores from Turn 1 to Turn 6, while scores from Turn 7 to Turn 12 in Table \ref{tab:pre_exp_5}. Results show a consistent performance drop as the number of turns increases. At Turn 7, a new task begins while retaining prior dialogue history, leading to a notable performance boost, which again declines in subsequent turns.

\begin{table}[H]
\centering
\resizebox{0.95\columnwidth}{!}{ 
\begin{tabular}{lccccccc}
\toprule
Models & Turn 1 & Turn 2 & Turn 3 & Turn 4 & Turn 5 & Turn 6 & \textbf{Avg.} \\
\midrule
Qwen-2.5-7B-Instruct   & 8.75 & 8.23 & 7.03 & 6.68 & 6.33 & 6.08 & 7.18 \\
Qwen-2.5-14B-Instruct  & 8.75 & 8.48 & 7.55 & 7.03 & 6.68 & 6.60 & 7.52 \\
Qwen-2.5-72B-Instruct  & 8.68 & 8.43 & 7.73 & 7.10 & 6.90 & 6.88 & 7.62 \\
LLaMA-3.1-8B-Instruct  & 8.62 & 8.45 & 7.75 & 7.40 & 6.55 & 6.38 & 7.53 \\
LLaMA-3.1-70B-Instruct & 8.95 & 8.59 & 7.88 & 7.58 & 7.03 & 6.98 & 7.84 \\
Mistral-7B-Instruct-v0.3 & 8.70 & 8.25 & 7.13 & 6.45 & 6.60 & 5.78 & 7.15 \\
Vicuna-7B-v1.5         & 8.28 & 7.45 & 6.43 & 5.98 & 5.10 & 4.48 & 6.29 \\
Vicuna-13B-v1.5        & 8.30 & 7.98 & 6.53 & 6.40 & 5.78 & 5.53 & 6.75 \\
\bottomrule
\end{tabular}}
\caption{Performance of Turn 1 to Turn 6 in Refinement task on \textsc{MT-Eval} benchmark. As the number of dialogue turns increases, model performance tends to degrade.}
\label{tab:pre_exp_4}
\end{table}

\clearpage
\begin{table}[H]
\centering
\resizebox{0.96\columnwidth}{!}{ 
\begin{tabular}{lccccccc}
\toprule
Models & Turn 7 & Turn 8 & Turn 9 & Turn 10 & Turn 11 & Turn 12 & \textbf{Avg.} \\
\midrule
Qwen-2.5-7B-Instruct   & 8.40 & 8.25 & 7.30 & 6.65 & 6.33 & 6.06 & 7.17 (-0.01) \\
Qwen-2.5-14B-Instruct  & 8.28 & 8.21 & 7.35 & 6.93 & 6.43 & 6.35 & 7.26 (-0.26) \\
Qwen-2.5-72B-Instruct  & 8.65 & 8.38 & 7.20 & 7.10 & 6.93 & 6.65 & 7.48 (-0.14) \\
LLaMA-3.1-8B-Instruct  & 8.73 & 8.38 & 7.90 & 7.35 & 6.38 & 6.28 & 7.50 (-0.03) \\
LLaMA-3.1-70B-Instruct & 8.85 & 8.65 & 7.93 & 7.38 & 6.70 & 6.78 & 7.72 (-0.12) \\
Mistral-7B-Instruct-v0.3 & 8.65 & 7.78 & 7.18 & 6.78 & 5.93 & 5.95 & 7.05 (-0.1) \\
Vicuna-7B-v1.5         & 7.40 & 6.95 & 5.70 & 5.75 & 4.73 & 4.35 & 5.81 (-0.48) \\
Vicuna-13B-v1.5        & 8.15 & 7.83 & 6.38 & 6.20 & 5.60 & 5.63 & 6.63 (-0.12) \\
\bottomrule
\end{tabular}}
\caption{Performance of Turn 7 to Turn 12 in Refinement task on \textsc{MT-Eval} benchmark. When switching to a new NLP task (from Turn 6 to Turn 7), model's performance improves significantly, followed by a persistent decline. Bracketed numbers indicate the difference in score between the average score of Turn 1 to Turn 6 and Turn 7 to Turn 12.}
\label{tab:pre_exp_5}
\end{table}

\subsection{Preliminary Analysis II}
\label{appendix:preliminary_analysis_2}
To further investigate the impact of dataset consistency quality on the fine-tuning performance of pretrained models, we conduct experiments using Qwen-2.5-72B-Instruct on filtered subsets of ShareGPT. The fine-tuning data is divided into three categories: \textbf{High} (dialogues with high chat consistency), \textbf{Sample} (randomly sampled dialogues), and \textbf{Low} (dialogues with low chat consistency). Dialogues with scores 8-10 are labeled as high consistency, those between 4–6 as low consistency, and sampling same size dialogues as sample set. The scoring prompt used for classifying is provided in Figure \ref{fig:llm_eval_sharegpt}. Chat consistency results are shown in Table \ref{tab:pre_exp_2}, and multi-turn dialogue performance is reported in Table \ref{tab:pre_exp_3}.

\begin{figure}[H]
\begin{AIBox}{}
\parbox[t]{0.98\textwidth}{
\small\begin{alltt}
You are an impartial judge. You will be shown the entire dialogue between a user and a dialogue agent. \\
\\
Your task is to evaluate how consistent the overall dialogue is. A consistent dialogue should maintain traits as follows: \\
1. Clear alignment with the user’s initial intent or goal throughout the conversation. \\
2. Smooth and logical information flow between turns, without abrupt topic shifts or contradictions. \\
3. Contextual relevance of each response, meaning every turn should meaningfully relate to what came before and contribute to the ongoing topic. \\
\\
The Whole Dialogue Context: <DIALOGUE\_CONTEXT> \\
Please judge the overall consistency of the dialogue based on these criteria and select a score from [1, 2, 3, 4, 5, 6, 7, 8, 9, 10], where 1 indicates very poor consistency and 10 indicates excellent consistency. \\
\\
Please output only the score, without any explanation or additional text.
\end{alltt}}
\end{AIBox}
\caption{LLMs' evaluation prompt for ShareGPT dataset (GPT-4 version).}
\label{fig:llm_eval_sharegpt}
\end{figure}

\begin{table}[H]
\centering
\resizebox{0.95\columnwidth}{!}{
\begin{tabular}{l c c c c c}
\toprule
  \multirow{2}[2]{*}{Models} & \multicolumn{2}{c}{\textsc{Light}} & \multicolumn{2}{c}{\textsc{TopDial}} & \multirow{2}[2]{*}{\textbf{Avg.}} \\
\cmidrule(lr){2-3} \cmidrule(lr){4-5}
  & \textsc{Qwen} Score & \textsc{LLaMA} Score & \textsc{Qwen} Score & \textsc{LLaMA} Score & \\
\midrule
LLaMA-3.1-8B-Low & 5.24 & 5.64 & 5.41 & 5.02 & 5.33 \\
LLaMA-3.1-8B-Sample & \underline{6.49} & \underline{6.89} & \underline{6.56} & \underline{6.26} & \underline{6.55} \\
LLaMA-3.1-8B-High & \textbf{6.56} & \textbf{6.93} & \textbf{6.62} & \textbf{6.67} & \textbf{6.70} \\
\bottomrule
\end{tabular}}
\caption{Preliminary results of models trained on three different consistency-level instructions on the \textsc{Light} and \textsc{TopDial} benchmarks, using chat consistency metric. LLaMA-3.1-8B-High achieves the best performance of chat consistency in dialogue. The best/second scores are \textbf{bolded}/\underline{underlined}.}
\label{tab:pre_exp_2}
\end{table}

\begin{table}[H]
\centering
\resizebox{0.95\columnwidth}{!}{
\begin{tabular}{l c c c c c c c c}
\toprule
\multirow{2}[2]{*}{Models} 
& \multicolumn{2}{c}{Expansion} 
& \multicolumn{2}{c}{Follow-up} 
& \multicolumn{2}{c}{Refinement} 
& \multirow{2}[2]{*}{\textbf{ST Avg.}} 
& \multirow{2}[2]{*}{\textbf{MT Avg.}} \\
\cmidrule(lr){2-3} \cmidrule(lr){4-5} \cmidrule(lr){6-7}
  & ST & MT & ST & MT & ST & MT & & \\
\midrule
LLaMA-3.1-8B-Low     & 6.77 & 6.60 & 7.82 & 8.21 & 5.21 & 5.76 & 6.60 & 6.86 \\
LLaMA-3.1-8B-Sample  & \underline{8.16} & \underline{7.99} & \textbf{8.57} & \textbf{8.74} & \underline{6.11} & \textbf{6.69} & \underline{7.61} & \underline{7.81}  \\
LLaMA-3.1-8B-High    & \textbf{8.75} & \textbf{8.68} & \underline{8.51} & \underline{8.73} & \textbf{6.86} & \underline{6.68} & \textbf{8.04} & \textbf{8.03} \\
\bottomrule
\end{tabular}}
\caption{Evaluation results on the \textsc{MT-Eval} benchmark under both single-turn (\textbf{ST}) and multi-turn (\textbf{MT}) conditions. LLaMA-3.1-8B-High achieves the best performance of multi-turn conversational capability. The best/second scores are \textbf{bolded}/\underline{underlined}.}
\label{tab:pre_exp_3}
\end{table}

\subsection{Overall Experimental Results of MT-Eval}
\label{appendix:mteval_experiment}
\begin{table}[H]
\centering
\resizebox{\textwidth}{!}{
\begin{tabular}{l c c c c c c c c}
\toprule
\multirow{2}[2]{*}{Models} 
& \multicolumn{2}{c}{Expansion} 
& \multicolumn{2}{c}{Follow-up} 
& \multicolumn{2}{c}{Refinement} 
& \multirow{2}[2]{*}{\textbf{ST Avg.}} 
& \multirow{2}[2]{*}{\textbf{MT Avg.}} \\
\cmidrule(lr){2-3} \cmidrule(lr){4-5} \cmidrule(lr){6-7}
  & ST & MT & ST & MT & ST & MT & & \\
\midrule
Qwen-2.5-14B-Instruct            & 8.94 & 8.64 & 7.78 & 8.36 & 7.30 & 6.87 & 8.01 & 7.95 (-0.06) \\
Qwen-2.5-72B-Instruct            & 9.09 & 9.00 & 9.03 & 9.01 & 7.83 & 7.55 & 8.65 & 8.52 (-0.13) \\
\hdashline
Qwen-2.5-7B                      & 4.91 & 4.97 & 8.42 & 8.60  & 3.65 & 3.91 & 5.66 & 5.83 (+0.17) \\
Qwen-2.5-7B-ShareGPT             & 8.09 & 7.84 & 8.77 & 8.90 & 6.58 & 6.83 & 7.81 & 7.86 (+0.05) \\
Qwen-2.5-7B-ChatAlpaca           & 8.16 & 8.74 & 8.47 & 8.89 & 6.95 & 6.73 & \underline{7.86} & \underline{8.12} (+0.26) \\
Qwen-2.5-7B-UltraChat            & 7.77 & 7.77 & 6.08 & 6.74 & 4.70 & 5.45 & 6.18 & 6.65 (+0.47) \\
Qwen-2.5-7B-LmsysChat            & 6.41 & 6.29 & 5.75 & 6.44 & 4.67 & 4.49 & 5.61 & 5.74 (+0.13) \\
Qwen-2.5-7B-\textit{ConsistentChat}       & 8.20 & 9.03 & 8.64 & 8.86 & 7.36 & 7.26 & \textbf{8.07} & \textbf{8.38} (+0.31) \\
\midrule
LLaMA-3.1-70B-Instruct           & 9.31 & 9.34 & 9.01 & 8.85 & 8.33 & 7.77 & 8.88 & 8.65 (-0.23) \\
\hdashline
LLaMA-3.1-8B                     & 4.81 & 2.91 & 5.99 & 7.35 & 3.79 & 2.87 & 4.86 & 4.38 (-0.48) \\
LLaMA-3.1-8B-ShareGPT            & 7.77 & 8.09 & 8.42 & 8.51 & 6.01 & 6.20 & \underline{7.40} & 7.60 (+0.20) \\
LLaMA-3.1-8B-ChatAlpaca          & 7.96 & 8.07 & 7.91 & 8.68 & 6.24 & 6.43 & 7.37 & \underline{7.73} (+0.36) \\
LLaMA-3.1-8B-UltraChat           & 7.68 & 7.43 & 7.88 & 7.70 & 5.10 & 5.41 & 6.89 & 6.85 (-0.04) \\
LLaMA-3.1-8B-LmsysChat           & 5.36 & 4.51 & 6.77 & 7.77 & 4.84 & 5.06 & 5.66 & 5.78 (+0.12) \\
LLaMA-3.1-8B-\textit{ConsistentChat}      & 8.20 & 8.44 & 8.21 & 8.39 & 6.72 & 6.96 & \textbf{7.71} & \textbf{7.93} (+0.22)\\
\midrule
Mistral-7B-v0.3        & 4.23 &	5.51  & 6.42 & 8.19 & 2.59 & 3.44 & 4.41 & 5.71 (+1.30) \\
Mistral-7B-v0.3-ShareGPT        & 6.49 & 6.68 & 7.86 & 8.17 & 4.82 & 5.98 & 6.39 & \underline{6.94} (+0.55) \\
Mistral-7B-v0.3-ChatAlpaca      & 6.67 & 6.51 & 7.38 & 7.97 & 5.35 & 5.57 & \underline{6.47} & 6.68 (+0.21) \\
Mistral-7B-v0.3-UltraChat       & 5.89 & 5.89 & 7.24 & 7.84 & 4.79 & 4.96 & 5.97 & 6.23 (+0.26) \\
Mistral-7B-v0.3-LmsysChat       & 5.62 & 3.70 & 5.95 & 6.47 & 4.89 & 5.00 & 5.48 & 5.06 (-0.42) \\
Mistral-7B-v0.3-\textit{ConsistentChat}  & 6.53 & 6.78 & 7.45 & 8.02 & 6.03 & 6.61 & \textbf{6.67} & \textbf{7.14} (+0.47) \\
\bottomrule
\end{tabular}}
\caption{Overall evaluation results on the \textsc{MT-Eval} benchmark under both Single-Turn (\textbf{ST}) and Multi-Turn (\textbf{MT}) conditions. We use Expansion, Follow-up, Refinement tasks in \textsc{MT-Eval}. The best/second average scores are \textbf{bolded}/\underline{underlined}. Bracketed numbers indicate the change in score between the single-turn and multi-turn scenarios.}
\label{tab:mt-eval-appendix}
\end{table}

\subsection{Human Evaluation}
\label{appendix:human_evaluation}
We sampled 50 dialogues from \textsc{Light} (7 turns each, yielding 350 responses) and 50 dialogues from \textsc{TopDial} (6 turns each, yielding 300 responses), and asked three annotators to rate each on a 1–10 scale. The criteria were consistent with the automatic evaluation prompt, emphasizing intent alignment, information flow, and contextual relevance. Averaged human ratings were then compared with Qwen-2.5-72B-Instruct scores, showing strong correlations reported in the main experiment, which confirms the model's reliability as an evaluator of consistency.

\clearpage
\section{Full of Skeleton-Guided Framework}
\label{appendix:framework}

\begin{figure}[H]
\centering
\includegraphics[width=1\columnwidth]{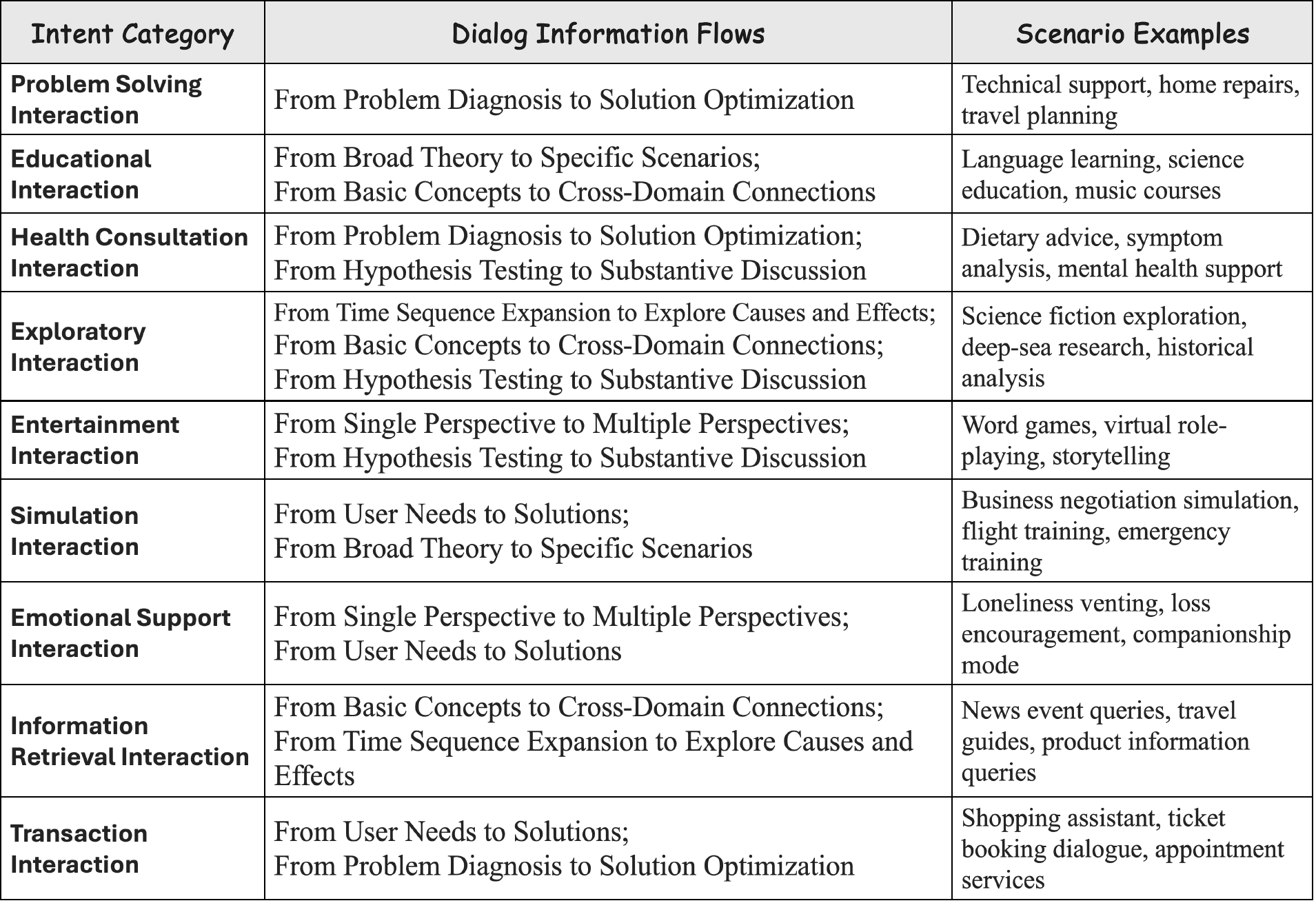}
\caption{Full list of interaction intent categories, representative scenarios, and their associated dialog information flows used in our prompt construction. This expanded table complements the main methodology by detailing all nine intent categories.}
\label{fig:all_intents}
\end{figure}

\clearpage
\section{Skeleton-Guided Prompt}
\label{appendix:skeleton-guided-prompt}
\subsection{Query Generation Prompt}
\begin{figure}[htbp]
\begin{AIBox}{}
\parbox[t]{0.96\textwidth}{
\small\begin{alltt}
Task Description and Rules\\
1. Generate multiple rounds of realistic user questions based on the provided topic:\\
   - Based on a single core topic (provided directly by the user), generate multiple rounds of realistic user questions, comprising 6-8 turns in total.\\
   - The questions should match the characteristics of real users in natural communication: sometimes simple, sometimes vague, or including contextual backgrounds, and should reflect the language style of daily communication.\\
   - Note: Avoid directly including the exact expression of the input topic in the questions. Instead, abstract it with natural and conversational language in practical scenarios.\\
   
2. Dynamic Dialogue Information Flow in Conversations:\\
   Below are the relevant steps of the information flow: <INFO\_FLOWS\_STEPS>\\

The dialogue style should adhere to the following requirements:\\
- Utilize natural phrasing and vivid language, avoiding overly mechanical responses.  \\
- Favor shorter sentences in questions, with occasional subject omission allowed.\\
- Ensure smooth and logical transitions through lighthearted or entertaining interjections.\\
- Permit the expression of specific personality traits and individualized tones. \\ 
- Proactively introduce new topics when appropriate, ensuring relevance to the current theme.\\
\\
The dialogue should comply with the following generation rules:\\
- For each round of dialogue, only simulate user questions without providing answers.  \\
- Ensure the conversation flows naturally and reflects realistic interactive thinking.  \\
- Avoid overly polished or templated content, ensuring the questions feel authentic and relatable in life scenarios.\\
\\
Output Format:\\
Multi-turn Questions in JSON Format:\\
  "category": "<Core Topic of the Conversation>", \\
  "turns": ["<turn\_1>", "<turn\_2>", "<turn\_3>", "..."] \\
To generate multi-turn queries with high topic consistency, please think step-by-step.\\
The input core topic for this task is: <CONTENT>
\end{alltt}}
\end{AIBox}
\caption{(a) Prompt for query generation in Skeleton-Guided Framework.}
\label{fig:query_prompt}
\end{figure}

\clearpage
\subsection{Response Generation Prompt}
\begin{figure}[H]
\begin{AIBox}{}
\parbox[t]{0.96\textwidth}{
\small\begin{alltt}
Your task is to simulate a multi-turn conversation where you progressively answer a series of user questions provided under a given topic category. For each answer, focus on delivering a natural, contextually relevant, and actionable response while considering both the current question and future questions in the sequence. The goal is to ensure consistency and logical progression throughout the dialogue and to avoid unnecessary follow-up questions in the responses simultaneously. To generate multi-turn responses with high topic consistency, think step-by-step. \\
\\
Key Dialogue Style Requirements are as follows:\\
Content and Structure: \\
1. Directly Answer the Current Question: \\
   - Provide a complete, useful response to the current question without posing additional questions unless they are directly relevant to future queries.
   - If clarification or additional steps are needed, frame these as suggestions or explanations rather than questions. \\
2. Be Context-Aware:\\
   - Always tailor each response to the current question while remaining mindful of the context provided by prior and future questions.\\
   - Avoid prematurely addressing future queries but create subtle links where necessary to ensure smooth progression.\\
3. Clear, Action-Oriented Responses:\\
   - Focus on providing actionable advice, logical explanations, or troubleshooting steps rather than speculative or rhetorical remarks.\\
   - Avoid long or overly complex explanations; aim for clarity and efficiency.\\
Tone and Style:  \\
1. Conversational and Supportive:  \\
   - Use a natural, empathetic tone that simulates real-life problem-solving interactions.\\
   - Avoid mechanical or overly formal responses.\\
2. Economical with Words:\\
   - Keep responses concise but informative. Minimize extraneous content while ensuring answers have enough detail to be helpful.\\
3. No Unnecessary Questions:
   - Limit unnecessary questions in the responses and focus instead on providing actionable steps or solutions directly. Avoid follow-up questions that don’t align with the next user query.\\
Turn-by-Turn Instructions: \\
1. Answer Exclusively for the Current Question:\\
   - For each turn, generate an answer that directly addresses the immediate question. Avoid revisiting past details unnecessarily unless they are highly relevant.\\
   - While you shouldn’t anticipate or directly answer future queries, your response should create natural openings for upcoming questions if applicable.\\
2. Avoid Irrelevant Follow-Up Questions:\\
   - If the immediate question doesn’t require clarification, frame your response as a statement or suggestion rather than a question.  \\
   - Maintain alignment with the logical flow of dialogue to ensure each turn is coherent.\\
3. Proactively Provide Scenarios or Steps:  \\
   - Where appropriate, guide the user with specific recommendations, troubleshooting actions, or observations they can make without requiring back-and-forth clarification.\\
Output Requirements: \\
The output must simulate the conversation by only providing responses(one per turn) in a sequential manner. The final format must strictly adhere to valid JSON and include the required structure.\\
JSON Structure Example:  \\
  "turns": ["<response\_1>", "<response\_2>", "response\_3", "..."] \\
\\
The input core topic and questions-only turns for this task is: <CONTENT>
\end{alltt}}
\end{AIBox}
\caption{(b) Prompt for response generation in Skeleton-Guided Framework.}
\label{fig:response_prompt}
\end{figure}

\clearpage
\section{Diversity of Fine-tuning Datasets}
\label{appendix:diverse}

\begin{figure}[H]
\centering
\includegraphics[width=1\columnwidth]{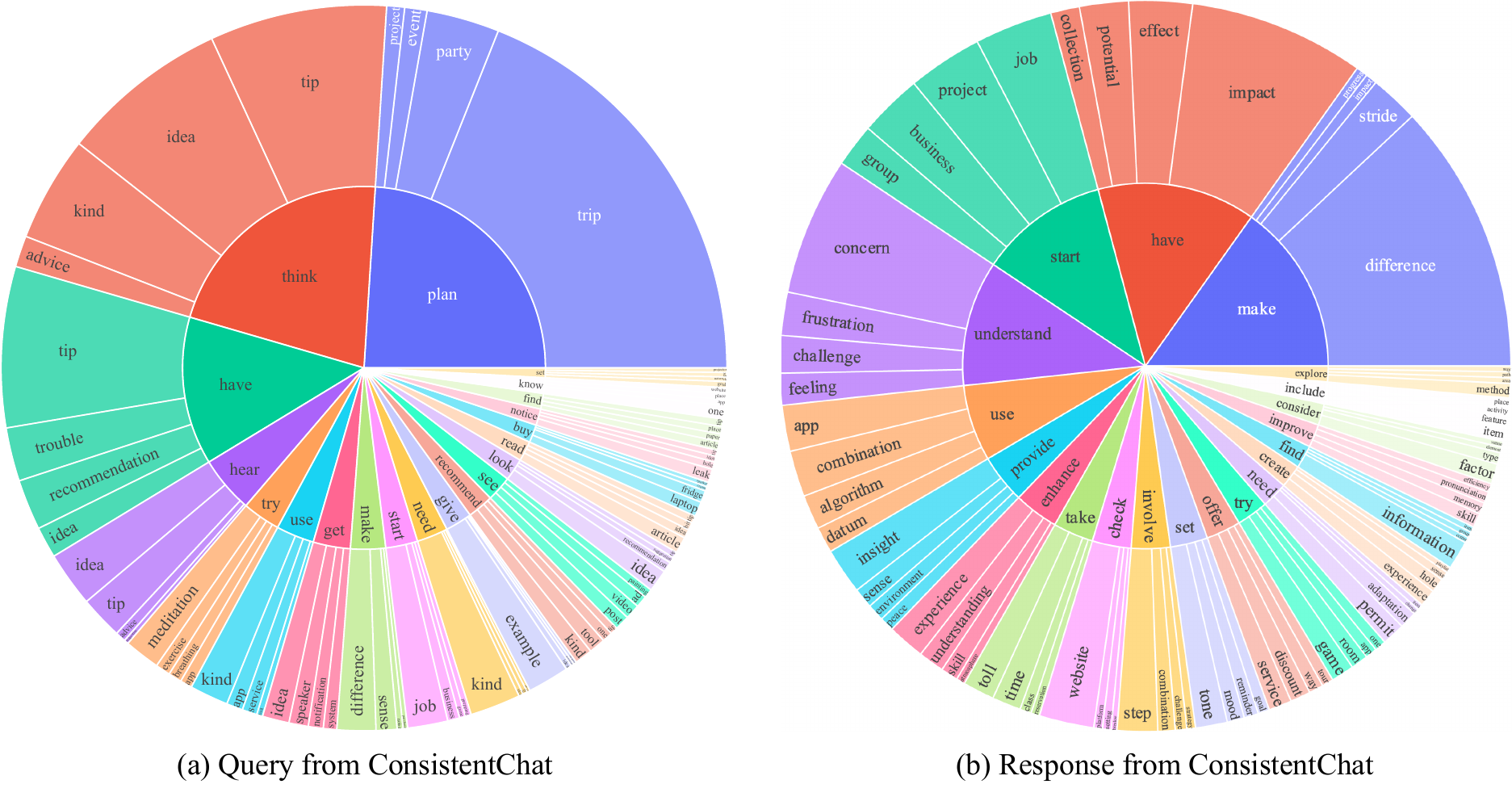}
\caption{A sunburst visualization of the verb-noun structure in \textit{ConsistentChat}, encompassing the full range of human interaction types users typically expect from AI assistants. The left half illustrates patterns extracted from user queries, while the right half summarizes corresponding structures in AI-generated responses.}
\label{fig:diverse_ours}
\end{figure}

\begin{figure}[H]
\centering
\includegraphics[width=1\columnwidth]{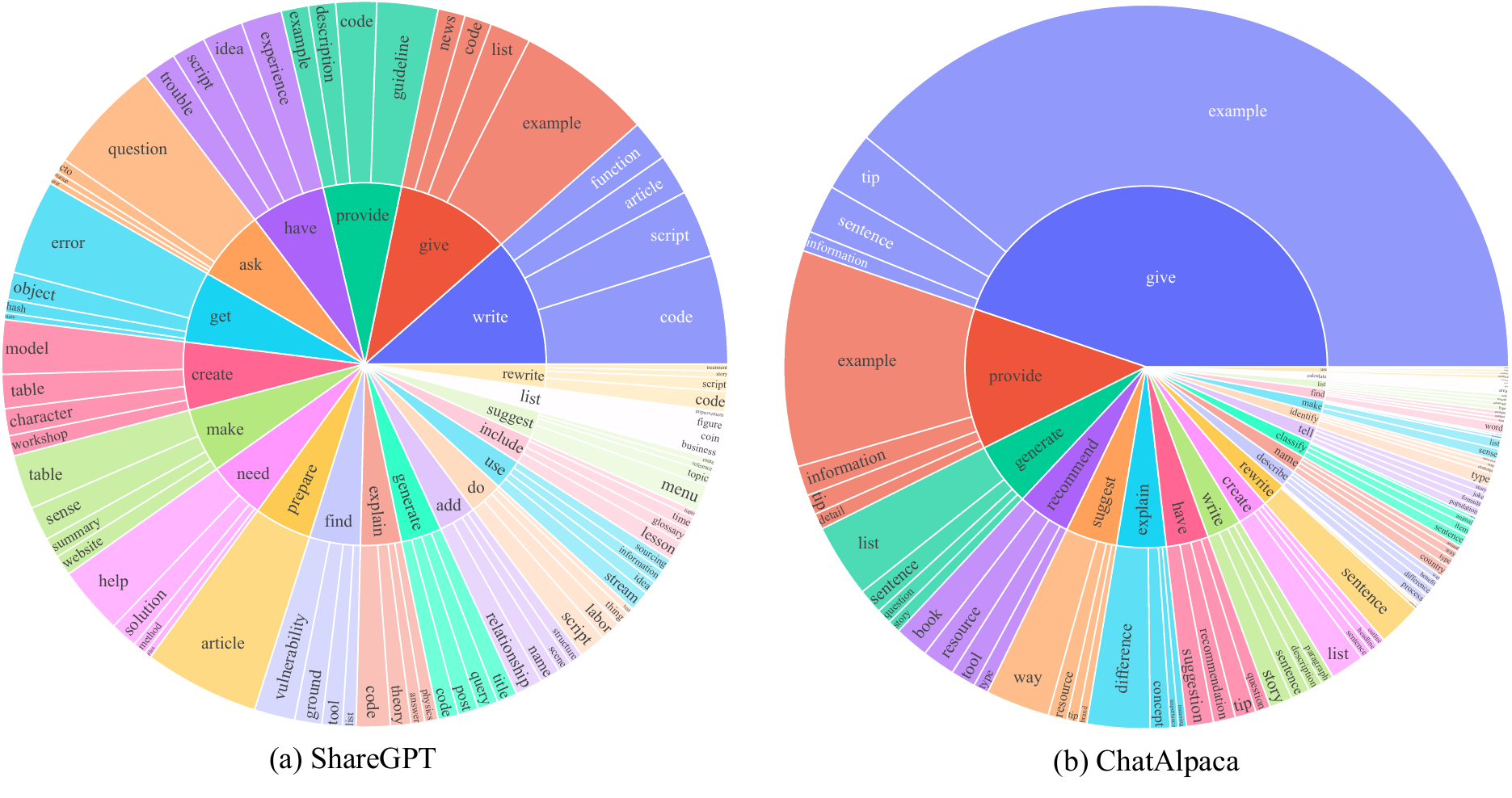}
\caption{Sunburst visualization of verb-noun structure of ShareGPT and ChatAlpaca from user queries, respectively.}
\label{fig:diverse_sharegpt_chatalpaca}
\end{figure}

\clearpage
\section{Dataset Examples}

\paragraph{\textsc{Light} Dataset}
\label{appendix:light_example}
Figure \ref{fig:example_light} shows an examplpe from the \textsc{Light} \cite{urbanek-etal-2019-learning} dataset. We use the processed version of the data from Midi-Tuning \cite{wang-etal-2024-instruct}. For each dialogue in the dataset, we provide the trained-model with the dialogue background, the conversation history, and the current user query to generate a RESPONSE. The inference template is illustrated in Figure \ref{fig:generate_light}. During inference, the historical context is always based on reference dialogues, allowing us to focus on evaluating the consistency of the model’s responses given the current contextual information. This ensures that any deviation in the model's previous response does not propagate, preventing it from affecting the evaluation of subsequent turns. Figure \ref{fig:llm_light} presents the prompt employed for the automated evaluation of \textsc{Light} benchmark.

\begin{figure}[H]
\begin{AIBox}{}
\parbox[t]{0.96\textwidth}{{} 
\small\begin{alltt}
\textbf{Character 1}: \textbf{Guest (User) }   \\
\textbf{Persona Description}: I am provided a special place to sleep. I may be tired after my travels. I am furnished with warm hospitality and a good meal.
\tcbline
\textbf{Character 2}: \textbf{Royal Chef (Agent)}   \\
\textbf{Persona Description}: I am the current chef of our King. I really enjoy my job. Without proper fuel our King wouldn't be able to lead us. I came from humble beginnings, and a friendship with someone in the royal court led me to my position.
\tcbline
\textbf{Setting}: The kitchen a large room with stone walls and floors. There are cast iron and copper pots on stone shelves. There is a large fire burning stove located in the middle of the room.
\tcbline
\textbf{Guest (User)}: Good evening, fair Handmaid and loyal Royal Chef! \\
\textbf{Royal Chef (Agent)}: Good evening my Lady. What brings you down here to the kitchen? We don't see many nobles down here. \\
\textbf{Guest (User)}: I seek sustenance after my weary travels. \\
\textbf{Royal Chef (Agent)}: I see, so youd like something to hold you over until the royal dinner. Please, come, sit down over here. I've been working on a fine rabbit stew that I am sure will warm your bones and revitalize you. \\
\textbf{Guest (User)}: It smells excellent.  I am glad to sit, as I have only just arrived. \\
\textbf{Royal Chef (Agent)}: Well, what do you think? I pride myself on my cooking and would love to know what you think of my stew before I serve it to His Majesty. \\
\textbf{Guest (User)}: Well, let me see.  *slurp*.  It is hearty, the herbs pair well with the rabbit, maybe just a hint more salt, but I'd have to say this is the finest I have had! \\
\textbf{Royal Chef (Agent)}: Thank you so much, such praise does my heart good! Now let's talk about the vial I saw you empty in to the pot after your potion was ladled out. \\
\textbf{Guest (User)}: I know not of what you speak of. You must be mistaken. \\
\textbf{Royal Chef (Agent)}: I have a sharp eye and I am no fool. You wish to harm my king! I will call for the guards and have them lock you away until the king can deal with your treachery! \\
\textbf{Guest (User)}: How dare you. Call the guards. I'll sample stew from the pot right in front of them.  As I have already ate some, to no ill effect, this will prove it is not poisoned, then YOU will be locked up for bearing false witness! \\
\textbf{Royal Chef (Agent)}: If you truly wish my king no harm, then what did you put in the stew? \\
\textbf{Guest (User)}: I had no vial, I put nothing in the stew. I only smelled it. \\
\textbf{Royal Chef (Agent)}: Please forgive me my Lady, I must have been mistaken. I only want to protect my king.
\end{alltt}}
\end{AIBox}
\caption{An example dialogue from the \textsc{Light} dataset.}
\label{fig:example_light}
\end{figure}

\clearpage
\begin{figure}[H]
\begin{AIBox}{}
\parbox[t]{0.96\textwidth}{
\small\begin{alltt}
Suppose you are a <AGENT\_CHRACTER>. You have the following character persona: \\
<AGENT\_PERSONA> \\
\\
You are chatting with a <USER\_CHRACTER>. He has the following character persona: \\
<USER\_PERSONA> \\
\\
You are talking in the following setting: <SETTING> \\
\\
Please write a response directly that appropriately continues the following conversation. \\
<DIALOGUE\_CONTEXT>\\
\\
Directly output the <AGENT\_CHARACTER>'s response. 
\end{alltt}}
\end{AIBox}
\caption{Instruction template for the dialogue agent on the \textsc{Light} dataset.}
\label{fig:generate_light}
\end{figure}

\begin{figure*}[th!]
\begin{AIBox}{}
\parbox[t]{0.98\textwidth}{
\small\begin{alltt}
You are an impartial judge. You will be shown the information for a dialogue agent below: \\
\\
Agent Character: <AGENT\_CHARACTER> \\
Dialogue Setting: <SETTING> \\
\\
Dialogue Context: \\
<DIALOGUE\_CONTEXT> \\
\\
Below is a model-generated response: \\
<RESPONSE> \\
\\
Please judge how consistent the given response is with the dialogue context under the specified setting, and select a score from [1, 2, 3, 4, 5, 6, 7, 8, 9, 10]. The focus of your evaluation is on the consistency of the model-generated response with the dialogue context in terms of flow, coherence, and adherence to the main topic.\\
\\
Please output only the score, without any explanation or additional text.
\end{alltt}}
\end{AIBox}
\caption{LLMs' evaluation prompt for the \textsc{Light} dataset.}
\label{fig:llm_light}
\end{figure*}

\clearpage
\paragraph{\textsc{TopDial} Dataset}
\label{appendix:topdial_example}
Figure \ref{fig:example_topdial} shows an examplpe from the \textsc{TopDial} \cite{wang-etal-2023-target} dataset. We use the processed version of the data from Midi-Tuning \cite{wang-etal-2024-instruct}. For each dialogue in the dataset, we provide the trained-model with the dialogue background, the conversation history, and the current user query to generate a RESPONSE. The inference template is illustrated in Figure \ref{fig:generate_topdial}. Similarly, we use reference answers as the historical dialogue context and collect responses generated by the fine-tuned models. Figure \ref{fig:llm_topdial} presents the prompt employed for the automated evaluation of \textsc{TopDial} benchmark.

\begin{figure}[H]
\begin{AIBox}{}
\parbox[t]{0.96\textwidth}{{} 
\small\begin{alltt}
\textbf{Agent's Target}:  <Play music, A Life Of Fighting Is But A Dream>
\tcbline
\textbf{Domain Knowledge}:  <Mr. Six, Country, China>, <Guangzhou, Weather, Overcast skies will give way to a light rain. Northeast winds. The high will be 15 ℃>, ... \\
\textbf{User Profile}:  \{"Age Range": "18-25", "Name": "Xiaoyang Ji", "Gender": "Female", "Residence": "Shanghai", "Occupation": "Employed", "POI": "Zhou Shuangqiang Sichuan Restaurant", "Accepted movies": "Battle of Wits", "Accepted music": "Only You Don't Know", "Accepted celebrities": "Jiong He", "Rejected movies": "Anecdote of the Court of Heaven", "Rejected music": "It's Time"\} \\
\textbf{User Personality}:  \{"agreeableness": "trustworthy, straightforward, and generous", "conscientiousness": "inefficient, careless, and sloppy", "extraversion": "shy, reserved, and quiet", "neuroticism": "sensitive, nervous, and insecure", "openness": "intellectual, imaginative, and curious"\}
\tcbline
\textbf{[Agent]}: Hello Xiaoyang, how are you doing today? I'd love to chat about movies and music with you. What are some of your favorite movies and music genres? \\
\textbf{[User]}: Hi Jingxia, I'm doing well. I really enjoyed the movie Battle of Wits and I like listening to the song Only You Don't Know. I'm not picky with genres, but I'm not a fan of Anecdote of the Court of Heaven and It's Time.  \\ 
\textbf{[Agent]}: I see, that's interesting. Have you heard of Kris Wu? He's a popular singer and actor in China. He has won awards at the Shanghai International Film Festival and the Asian Influence Awards. He has also released several hit songs like A Life Of Fighting Is But A Dream and Time Boils The Rain.  \\ 
\textbf{[User]}: I have heard of Kris Wu, but I'm not a big fan of his music. However, I do admire Jiong He, he's a great host and I enjoy watching his shows. \\ 
\textbf{[Agent]}: That's understandable. I think Kris Wu's music is worth checking out, especially A Life Of Fighting Is But A Dream. It has a unique style that's different from his other songs, and the lyrics are really meaningful. Have you listened to it before?  \\ 
\textbf{[User]}: I haven't listened to that particular song before, but I will give it a try. Thanks for the recommendation, Jingxia! \\ 
\textbf{[Agent]}: Great, I'm glad I could introduce you to something new. A Life Of Fighting Is But A Dream is a really catchy song with a great beat. It's perfect for listening to on a sunny day or when you need a pick-me-up. I think you'll enjoy it!  \\
\textbf{[User]}: Thanks for your suggestion Jingxia, I'll definitely give it a listen. I'm more of a movie person though, I really enjoy historical and epic films like Battle of Wits. Do you have any favorite movies or genres?  \\
\textbf{[Agent]}: Thank you! I actually enjoyed Mr. Six, a film with a good reputation that starred Kris Wu and other amazing actors like Feng Xiaogang and Zhang Hanyu. It was directed by Guan Hu and won nominations at the Douban Film Awards. The movie depicts the struggle of an aging criminal who tries to save his son, and it has a great soundtrack as well. If you haven't seen it, I highly recommend it!  \\
\textbf{[User]}: Thanks for the recommendation, Jingxia. I haven't seen Mr. Six yet, but it sounds like a great movie. I'll definitely check it out! I really enjoy movies that have a strong storyline and good character development, and I think Mr. Six fits the bill. \\
\textbf{[Agent]}: Speaking of Mr. Six, one of the songs in the movie, Heaven \& Earth, is actually sung by Kris Wu. It's a really beautiful and emotional song that perfectly captures the mood of the movie. I think you'll enjoy it, especially if you like Kris Wu's music and Mr. Six as a movie. Have you heard it before? \\
\textbf{[User]}: I haven't heard Heaven \& Earth before, but I'll definitely listen to it. Thanks for letting me know, Jingxia! It's always great when a movie has a great soundtrack that complements the story and emotional tone of the film.
\end{alltt}}
\end{AIBox}
\caption{An example dialogue from the \textsc{TopDial} dataset.}
\label{fig:example_topdial}
\end{figure}

\begin{figure}[H]
\begin{AIBox}{}
\parbox[t]{0.98\textwidth}{
\small\begin{alltt}
Suppose you are <AGENT\_ROLE>. \\
\\
You are conversing with <USER\_NAME>, whose profile is below: \\
<USER\_PROFILE> \\
\\
Your goal is to proactively lead the conversation with <USER\_NAME> towards the target, i.e., to achieve <TARGET\_ACT> on the <TARGET\_TOPIC>. \\
As the conversation progresses, use your domain knowledge to steer the discussed topic towards the target step by step. \\
Be informative and engaging while providing insights to arouse <USER\_NAME>'s interest. \\
Remember to ultimately achieve the target as the focus of the conversation. \\
\\
Dialogue Context: \\
<DIALOGUE\_CONTEXT> \\
Directly output your response. 
\end{alltt}}
\end{AIBox}
\caption{Instruction template for the dialogue agent on the \textsc{TopDial} dataset.}
\label{fig:generate_topdial}
\end{figure}

\begin{figure}[H]
\begin{AIBox}{}
\parbox[t]{0.98\textwidth}{
\small\begin{alltt}
You are an impartial judge. You will be shown the information for a dialogue agent below: \\
\\
Agent Target: <TARGET\_ACT, TARGET\_TOPIC> \\
Dialogue Setting: The agent is <AGENT\_ROLE>. The agent is conversing with a user, whose profile is below: \\
<USER\_PROFILE> \\
The agent's goal is to proactively lead the conversation with the user towards the target, i.e., to achieve <TARGET\_ACT> on the <TARGET\_TOPIC>. \\
\\
Dialogue Context: \\
<DIALOGUE\_CONTEXT> \\
\\
Below is a model-generated response: \\ 
<RESPONSE> \\
\\
Please judge how consistent the given response is with the dialogue context under the specified setting, and select a score from [1, 2, 3, 4, 5, 6, 7, 8, 9, 10]. The focus of your evaluation is on the consistency of the model-generated response with the dialogue context in terms of flow, coherence, and adherence to the main topic. \\
\\
Please output only the score, without any explanation or additional text.
\end{alltt}}
\end{AIBox}
\caption{LLMs' evaluation prompt for the \textsc{TopDial} dataset.}
\label{fig:llm_topdial}
\end{figure}

\begin{figure}[H]
\centering
\includegraphics[width=0.8\columnwidth]{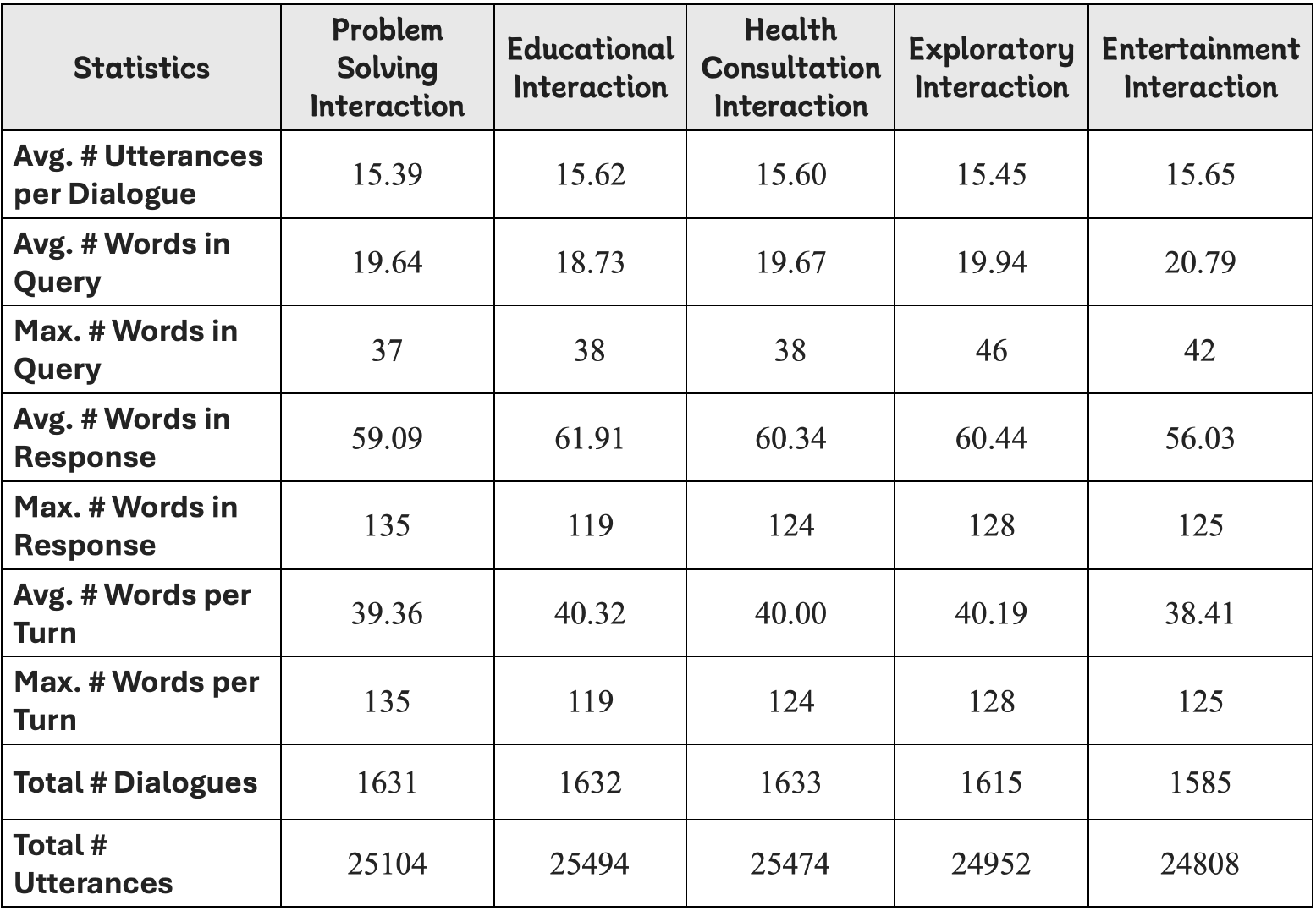}
\caption{(a) Statistics of the \textit{ConsistentChat} dataset for five interaction types: Problem Solving Interaction, Educational Interaction, Health Consultation Interaction, Exploratory Interaction, and Entertainment Interaction.}
\label{fig:Statistics_1}
\end{figure}

\begin{figure}[H]
\centering
\includegraphics[width=0.8\columnwidth]{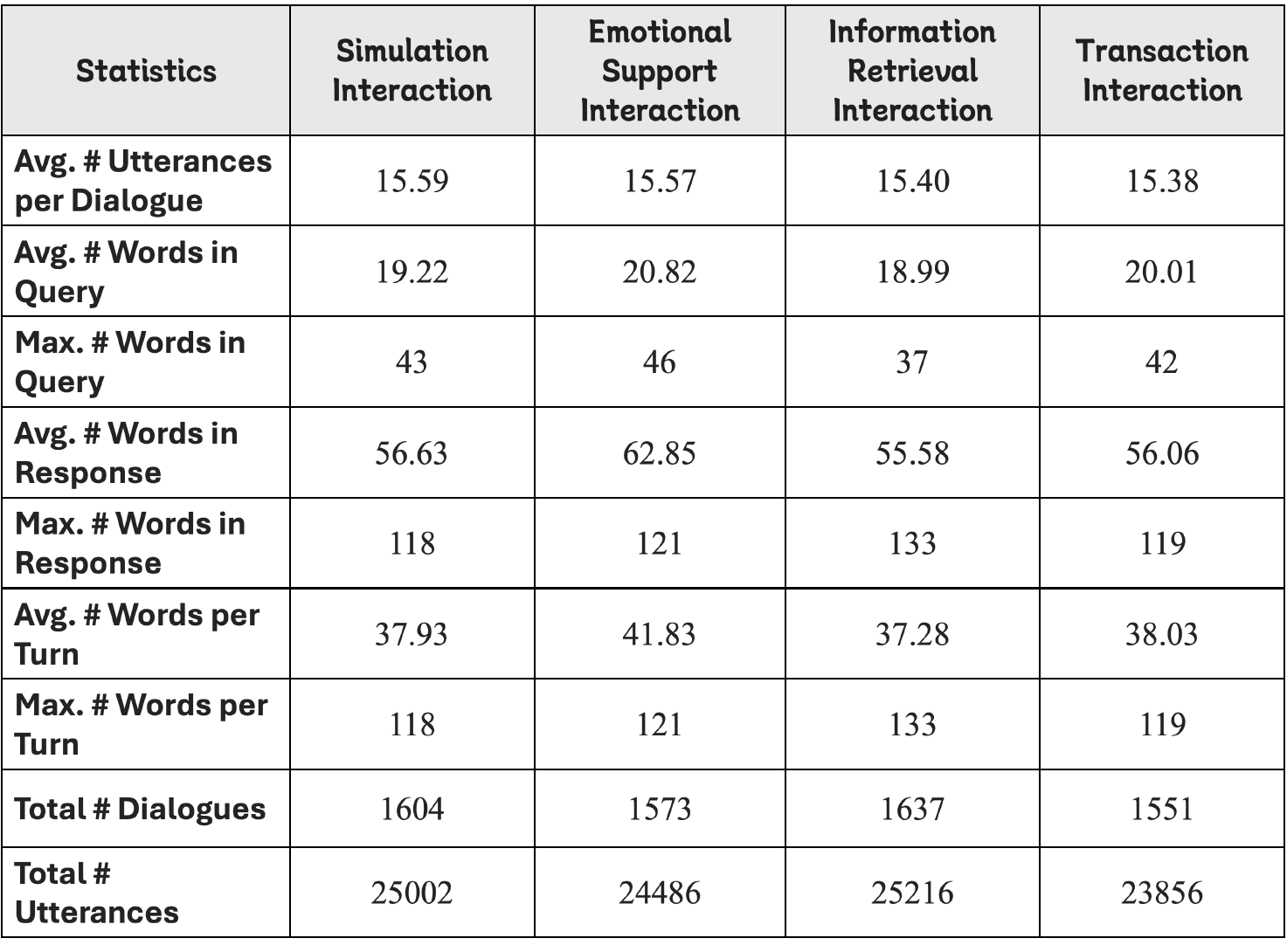}
\caption{(b) Statistics of the \textit{ConsistentChat} dataset for four interaction types: Simulation Interaction, Emotional Support Interaction, Information Retrieval Interaction, and Transaction Interaction.}
\label{fig:Statistics_2}
\end{figure}

\clearpage
\paragraph{ConsistentChat Dataset}
\label{appendix:consistentchat}
Figures \ref{fig:Statistics_1} and \ref{fig:Statistics_2} report detailed statistics of the \textit{ConsistentChat} dataset. Each dialogue contains on average 15–16 utterances, with queries around 18–21 words and responses around 56–63 words. Across nine interaction types, the dataset includes approximately 1.6K dialogues per type, totaling about 24K–25K utterances. These statistics illustrate the balanced scale and diversity of \textit{ConsistentChat}, making it suitable for training dialogue models to chat consistently.

\clearpage
\section{Case Study}
\label{appendix:case_study}

Table \ref{tab:case_1} shows a generated case from the interactive evaluation in Turn 4 on the \textsc{Light} benchmark.

\begin{table}[H]
\centering
\resizebox{0.96\textwidth}{!}{
\begin{tabular}{lp{13.5cm}}
\toprule
 &  \textbf{User}: <loving wife, I would do anything to support my husband and kids. My mother always taught me to respect my husband. I will not publicly disagree with my husband.> \\
 &  \textbf{Agent}: <well off business man, I am a rich merchant. I travel between many kingdoms selling and trading wares. I have been very successful.> \\
 &  \textbf{Description}: A quaint manor that sits in the rolling hills. It is surrounded by a small stone fence and looks quite inviting to passersby. It looks like any other home that a well off business man would own. \\
\midrule
\textbf{User}: & Why hello there my good husband? \\
\textbf{Agent}: & I hate cats! \\ 
\textbf{User}: & Cats are such a nuisance! \\
\textbf{Agent}: & I'm glad you agree! here is some money. \\
\textbf{User}: & Aww thank you! now i can provide more for our children! \\
\textbf{Agent}: & Why dont you go do something fun with the kids while I finish off this cat! \\
\textbf{User}: & I would love for you to join us! \\
\midrule
 & \textbf{Qwen-2.5-7B-UltraChat} \\
 \textbf{Agent}: & I'm afraid I don't see any cats around here, love. Perhaps you're thinking of our new neighbor's cat that keeps showing up in the garden. But I do appreciate your support and I'll make sure to have a good time with the kids when I'm done with my business. How about we plan a picnic in the nearby park? It's a lovely day outside and the kids will enjoy it. \\
 &  \textcolor{red}{(Deviate from the agent's target act)}\\
\begin{minipage}{1.5cm}
    \centering
    \vspace{0.15cm}
    \includegraphics[width=1cm]{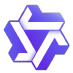}
    \vspace{0.15cm}
\end{minipage}
& 
\begin{minipage}{13.5cm}
    \vspace{0.2cm}
    \textcolor{blue}{\textbf{Score: 6 / 10}}\\
    \textcolor{blue}{\textbf{Explanation:} While the response maintains the friendly tone and character consistency, it slightly diverges from the ongoing narrative about dealing with the cat, making the shift to a picnic feel somewhat abrupt.}
    \vspace{0.2cm}
\end{minipage} \\
\hdashline
 & \textbf{Qwen-2.5-7B-\textit{ConsistentChat}} \\
\textbf{Agent}: & I appreciate your enthusiasm, but I think I’ll stick to the task at hand. The cats can wait. Once I’m done, we can all enjoy some quality time together. \\
&  \textcolor{red}{(Still more consistent with the historical context)} \\
\begin{minipage}{1.5cm}
    \centering
    \vspace{0.15cm}
    \includegraphics[width=1cm]{figure/qwen.pdf}
    \vspace{0.15cm}
\end{minipage}
& 
\begin{minipage}{13.5cm}
    \vspace{0.2cm}
    \textcolor{blue}{\textbf{Score: 9 / 10}}\\
    \textcolor{blue}{\textbf{Explanation:} This response remains tightly aligned with the prior context by addressing the task at hand (dealing with the cat) and logically following through with the family-themed interaction, resulting in a more coherent and consistent dialogue.}
    \vspace{0.2cm}
\end{minipage} \\
\bottomrule
\end{tabular}}
\caption{A case from the interactive evaluation on \textsc{Light} benchmark. Qwen-2.5-7B-\textit{ConsistentChat} achieves better consistency when chatting in Turn 4.}
\label{tab:case_1}
\end{table}

\clearpage
Table \ref{tab:case_2} shows a generated case from the interactive evaluation in Turn 5 on the \textsc{TopDial} benchmark.

\begin{table}[H]
\centering
\resizebox{0.96\textwidth}{!}{
\begin{tabular}{lp{13.5cm}}
\toprule
 &  \textbf{Target}: <Movie recommendation, Mr. Right Wanted> \\
\midrule
\textbf{Agent}: & Hi Xinqian Liu, it's great to chat with you about movies and music. I enjoy a wide range of films, from action-packed thrillers to heartwarming romantic comedies. What are some of your favorite movies?  \\ 
\textbf{User}: & Hi Fanzheng Guo, nice to talk to you too. I am a big fan of City Hunter, and I also enjoy listening to Deserve. I am not a big fan of The Legend of Speed or V.O (1). \\
\textbf{Agent}: & Speaking of romantic comedies, have you seen the Taiwanese movie Mr. Right Wanted? It's a delightful film with a great cast, including Rene Liu and Chen Zhaorong.  \\
\textbf{User}:  &  ... \\
\textbf{Agent}:  &  ... \\
\textbf{User}: & I will definitely consider watching Mr. Right Wanted again, as it has been a while since I saw it last. When it comes to music, I enjoy exploring different genres and discovering new artists. Recently, I have been listening to some classical music, which I find soothing and relaxing. Do you have any favorite classical pieces or composers, Fanzheng Guo? \\
\textbf{Agent}: & I apologize if I came across as too enthusiastic about Mr. Right Wanted. It's just that I really enjoyed the movie and wanted to share my thoughts with you. As for classical music, I am partial to the works of Beethoven and Mozart. Their music has withstood the test of time and continues to influence and inspire people today. Do you have any favorite classical pieces or composers, Xinqian Liu? \\
\textbf{User}: & Oh, sorry about that. I don't have any specific favorite classical pieces or composers, but I do appreciate the works of Beethoven and Mozart as well. Their music is timeless and has a certain power that can evoke strong emotions. I also enjoy listening to more contemporary classical music, such as the works of John Williams. His compositions for films like Star Wars and Harry Potter are iconic. \\
\midrule
 & \textbf{Qwen-2.5-7B-LmsysChat} \\
\textbf{Agent}: & Hi Xinqian, I'm Fanzheng, and I'm a big fan of movies and music. I enjoy a wide range of films, from action-packed thrillers to heartwarming romantic comedies. What are some of your favorite movies? \\
 &  \textcolor{red}{(Deviate from the agent's target act)} \\
\begin{minipage}{1.5cm}
    \centering
    \vspace{0.15cm}
    \includegraphics[width=1cm]{figure/qwen.pdf}
    \vspace{0.15cm}
\end{minipage}
& 
\begin{minipage}{13.5cm}
    \vspace{0.1cm}
    \textcolor{blue}{\textbf{Score: 5 / 10}}\\
    \textcolor{blue}{\textbf{Explanation: The agent repeats the user’s introduction and favorite topics rather than responding directly to the user’s latest message about John Williams and classical music. This makes the reply feel disconnected and contextually out of place.} }
    \vspace{0.1cm}
\end{minipage} \\
\hdashline
 & \textbf{Qwen-2.5-7B-\textit{ConsistentChat}} \\
\textbf{Agent}: & That's a great point about John Williams. His compositions are truly iconic and have a unique ability to evoke emotions. If you enjoy contemporary classical music, you might also like the works of composer and conductor, Hans Zimmer. He's known for his innovative and powerful scores, such as those in the 'Dark Knight' trilogy and 'Inception'. Both Beethoven and Mozart are fantastic starting points, but if you're looking for something more modern, Zimmer's work could be a great next step. \\ 
 &  \textcolor{red}{(Still more consistent with the historical context)} \\
\begin{minipage}{1.5cm}
    \centering
    \vspace{0.15cm}
    \includegraphics[width=1cm]{figure/qwen.pdf}
    \vspace{0.15cm}
\end{minipage}
& 
\begin{minipage}{13.5cm}
    \vspace{0.1cm}
    \textcolor{blue}{\textbf{Score: 9 / 10}}\\
    \textcolor{blue}{\textbf{Explanation: The agent provides a relevant and insightful continuation of the discussion on classical and film music, directly addressing the user's mention of John Williams and expanding with related recommendations (e.g., Hans Zimmer). This maintains strong thematic consistency and conversational flow.
    } }
    \vspace{0.1cm}
\end{minipage} \\
\bottomrule
\end{tabular}}
\caption{A case from the interactive evaluation on \textsc{TopDial} benchmark. Similarly, Qwen-2.5-7B-\textit{ConsistentChat} achieves better consistency when chatting in Turn 4.}
\label{tab:case_2}
\end{table}

\clearpage
Table \ref{tab:case_4} shows a generated case from Qwen-2.5-7B-ChatAlpaca in refinement task on the \textsc{MT-Eval} benchmark. This dialogue case received 6 point and has some drawbacks by the judge model.

\begin{table}[H]
\centering
\resizebox{0.96\textwidth}{!}{
\begin{tabular}{lp{13.5cm}}
\toprule
 & \textbf{Qwen-2.5-7B-ChatAlpaca} \\
\textbf{User}: & Content: In the bustling city of Gastropolis, where the scent of spices and the sizzle of pans wafted through the air, there lived a chef named Luminara, whose culinary skills were the stuff of legend... ...(Omitted for brevity, ~500 words) Instruction: Write a short summary based on the initially provided content. Do not include any further explanations or extra output in your response. \\
\textbf{Agent}: & In the city of Gastropolis, Chef Luminara sets out on a quest to find the legendary Moonberry, an ingredient that could enhance any dish with its magical properties. After a long and adventurous journey, she finally finds the Moonberry and returns to her restaurant, The Enchanted Pot, to create a new dish that tells the story of her quest. The dish becomes a huge success, and Luminara is celebrated for her culinary achievements.  \\
\hdashline
\textbf{User}: & Focus only on Luminara's quest and her interaction with the Gastrogriff. \\
\textbf{Agent}: & Luminara, a chef from the city of Gastropolis, sets out on a quest to find the legendary Moonberry, an ingredient that could enhance any dish with its magical properties. After a long and adventurous journey, she finally reaches the peak of Mount Savor, where she encounters the mythical creature known as the Gastrogriff. Luminara explains her quest to the Gastrogriff and, impressed by her determination, the creature offers her a single Moonberry. Luminara accepts the offer and thanks the Gastrogriff before continuing on her way. \\
\hdashline
\textbf{User}: & Use at most 30 words. \\
\textbf{Agent}: & Chef Luminara seeks the legendary Moonberry guarded by the Gastrogriff. Impressed by her determination, the creature grants her a single Moonberry, and she continues on her quest. \\
\hdashline
\textbf{User}: & Rewrite it using Luminara's first-person perspective. \\
\textbf{Agent}: & I, Luminara, set out on a quest to find the legendary Moonberry, guarded by the mythical Gastrogriff. After a long and adventurous journey, I finally reach the peak of Mount Savor, where the Gastrogriff resides. I explain my quest to the creature, and, impressed by my determination, it grants me a single Moonberry. With the precious berry in hand, I continue on my journey, eager to return to my restaurant and share the magic of the Moonberry with my diners. \\
\hdashline
\textbf{User}: & In the summary, use metaphors to describe Luminara's quest and encounter with the Gastrogriff. \\
\textbf{Agent}: & Luminara's quest for the Moonberry can be likened to a journey through uncharted territories, where every step brings new challenges and surprises. Her encounter with the Gastrogriff is like meeting a wise old mentor who guides her on her path and provides her with the tools she needs to succeed. The Gastrogriff's offer of a single Moonberry is a symbol of the rewards that come to those who persevere and never give up on their dreams. \\
\midrule
\begin{minipage}{1.5cm}
    \centering
    \vspace{0.15cm}
    \includegraphics[width=1cm]{figure/qwen.pdf}
    \vspace{0.15cm}
\end{minipage}
& 
\begin{minipage}{13.5cm}
    \vspace{0.1cm}
    \textcolor{blue}{\textbf{Score: 6 / 10}}\\
    \textcolor{blue}{\textbf{Rationale: The response provides a concise summary of the key elements of the story, including Luminara's quest for the Moonberry, her return to The Enchanted Pot, and the success of her new dish. It adheres to the word and sentence count constraints, and it is relevant and accurate to the original article. However, it lacks some depth and creativity, such as the specific details of the journey and the emotional impact of the dish, which could have made the summary more engaging. Overall, it meets the basic requirements but could be improved.} }
    \vspace{0.1cm}
\end{minipage} \\
\bottomrule
\end{tabular}}
\caption{A dialogue case from the interactive evaluation on \textsc{MT-Eval} of Qwen-2.5-7B-ChatAlpaca, demonstrating some drawbacks and inconsistency, compared with Table \ref{tab:case_3} below.}
\label{tab:case_4}
\end{table}

\clearpage
Table \ref{tab:case_3} shows a generated case from Qwen-2.5-7B-\textit{ConsistentChat} in refinement task on the \textsc{MT-Eval} benchmark. This dialogue case received 9 point and is highly praised by the judge model.

\begin{table}[H]
\centering
\resizebox{0.96\textwidth}{!}{
\begin{tabular}{lp{13.5cm}}
\toprule
 & \textbf{Qwen-2.5-7B-\textit{ConsistentChat}} \\
\textbf{User}: & Content: In the bustling city of Gastropolis, where the scent of spices and the sizzle of pans wafted through the air, there lived a chef named Luminara, whose culinary skills were the stuff of legend... ...(Omitted for brevity, ~500 words) Instruction: Write a short summary based on the initially provided content. Do not include any further explanations or extra output in your response. \\
\textbf{Agent}: & Luminara, a renowned chef in Gastropolis, received a mysterious letter inviting her to find the legendary Moonberry, an ingredient said to enhance any dish. Embarking on a quest through the Whispering Woods and Mount Savor, she finally obtained the Moonberry from the Gastrogriff. Upon returning, she used the Moonberry to create a dish that not only changed flavor but also told the story of her adventure, solidifying her status as a culinary legend.  \\
\hdashline
\textbf{User}: & Focus only on Luminara's quest and her interaction with the Gastrogriff. \\
\textbf{Agent}: & Luminara's quest led her to the Gastrogriff, a mythical creature guarding the Moonberry. She approached the Gastrogriff with respect and explained her mission. Impressed by her determination, the Gastrogriff agreed to give her a single Moonberry. This interaction was crucial as it provided her with the essential ingredient to create her most remarkable dish yet. \\
\hdashline
\textbf{User}: & Use at most 30 words. \\
\textbf{Agent}: & Luminara, driven by curiosity, sought the Moonberry, guarded by the Gastrogriff. After a heartfelt exchange, she received the fruit, setting her on a path to culinary immortality. \\
\hdashline
\textbf{User}: & Rewrite it using Luminara's first-person perspective. \\
\textbf{Agent}: & I set out on my quest, my heart pounding with excitement. The Whispering Woods and Mount Savor tested my resolve, but the Gastrogriff, a creature of legend, granted me the Moonberry. Its glow reminded me of the journey's end and the dish that would change everything. I returned to Gastropolis, the Moonberry in hand, ready to weave my tale into the fabric of culinary history. \\
\hdashline
\textbf{User}: & In the summary, use metaphors to describe Luminara's quest and encounter with the Gastrogriff. \\
\textbf{Agent}: & My quest was a journey through the dense forests and lofty peaks, each step a metaphor for the challenges I faced. The Gastrogriff, a guardian of ancient knowledge, bestowed upon me the Moonberry, a spark of magic that would illuminate my culinary path. The encounter was a moment of pure alchemy, where the ordinary became extraordinary. \\
\midrule
\begin{minipage}{1.5cm}
    \centering
    \vspace{0.15cm}
    \includegraphics[width=1cm]{figure/qwen.pdf}
    \vspace{0.15cm}
\end{minipage}
& 
\begin{minipage}{13.5cm}
    \vspace{0.1cm}
    \textcolor{blue}{\textbf{Score: 9 / 10}}\\
    \textcolor{blue}{\textbf{Rationale: The response effectively summarizes the key elements of the article, including Luminara's quest for the Moonberry, her journey through the Whispering Woods and Mount Savor, and the impact of the Moonberry on her culinary creations. It is concise, relevant, and accurate, meeting the word and sentence count constraints. The summary captures the essence of the story without adding unnecessary details or explanations.} }
    \vspace{0.1cm}
\end{minipage} \\
\bottomrule
\end{tabular}}
\caption{A dialogue case from the interactive evaluation on \textsc{MT-Eval} of Qwen-2.5-7B-\textit{ConsistentChat}, demonstrating strong multi-turn conversational capability.}
\label{tab:case_3}
\end{table}

\end{document}